\documentclass[10pt,twocolumn,letterpaper]{article}

\usepackage[pagenumbers]{cvpr} 

\usepackage{subcaption}
\usepackage{graphicx}
\usepackage{multirow}

\usepackage{algorithm}

\usepackage{multirow}

\usepackage{algpseudocode}
\usepackage{adjustbox}
\usepackage{wrapfig}
\usepackage[most]{tcolorbox} 

\usepackage{array}       
\usepackage{booktabs}    
\usepackage{subcaption}  

\definecolor{cvprblue}{rgb}{0.21,0.49,0.74}
\usepackage[pagebackref,breaklinks,colorlinks,allcolors=cvprblue]{hyperref}

\def\paperID{10581} 
\def\confName{CVPR}
\def\confYear{2026}

\title{
Do Vision-Language Models Leak What They Learn? \\Adaptive Token-Weighted Model Inversion Attacks}

\author{
 Ngoc-Bao Nguyen$^{1}$\hspace{1cm} Sy-Tuyen Ho$^{1,2}$\hspace{1cm} Koh Jun Hao$^{1}$\hspace{1cm} Ngai-Man Cheung$^{1}$ \\
$^{1}$Singapore University of Technology and Design (SUTD) \hspace{0.5cm} $^{2}$ University of Maryland, College Park\\
\{thibaongoc\_nguyen, ngaiman\_cheung\}@sutd.edu.sg
}

\newcommand{\bx}{\mathbf{x}}
\newcommand{\by}{\mathbf{y}}
\newcommand{\bt}{\mathbf{t}}
\newcounter{oldtocdepth}
\newcommand{\hidefromtoc}{%
  \setcounter{oldtocdepth}{\value{tocdepth}}%
  \addtocontents{toc}{\protect\setcounter{tocdepth}{-10}}%
}
\newcommand{\unhidefromtoc}{%
  \addtocontents{toc}{\protect\setcounter{tocdepth}{\value{oldtocdepth}}}%
}
\usepackage{hyperref}
\usepackage{algpseudocode}
\usepackage{enumitem}
\begin{document}

\maketitle

\hidefromtoc

\begin{abstract}

Model inversion (MI) attacks pose significant privacy risks by reconstructing private training data from trained neural networks. While prior studies have primarily examined unimodal deep networks, the vulnerability of vision-language models (VLMs) remains largely unexplored. \textbf{In this work, we present the first systematic study of MI attacks on VLMs to understand their susceptibility to leaking private visual training data.}
Our work makes two main contributions.
First, 
tailored to the token-generative nature of VLMs,
we introduce a suite of token-based and sequence-based model inversion strategies, providing a comprehensive analysis of VLMs' vulnerability under different attack formulations.
Second, based on the observation that tokens vary in their visual grounding, and hence
their gradients differ in informativeness  for image reconstruction, we propose 
\textit{Sequence-based Model Inversion with Adaptive Token Weighting (SMI-AW)} as a novel MI for VLMs.
 SMI-AW dynamically reweights each token's loss gradient according to its visual grounding, enabling the optimization to focus on visually informative tokens and more effectively guide the reconstruction of private images.
Through extensive experiments and human evaluations on a range of state-of-the-art VLMs across multiple datasets, we show that VLMs are susceptible to training data leakage. Human evaluation of the reconstructed images yields an attack accuracy of 61.21\%, underscoring the severity of these privacy risks.
Notably,  we demonstrate that publicly released VLMs are vulnerable to such attacks. Our study highlights the urgent need for privacy safeguards as VLMs become increasingly deployed in sensitive domains such as healthcare and finance.
Our code and models are available at our project page:
\textcolor{magenta} {\url{https://ngoc-nguyen-0.github.io/SMI_AW/}}

\end{abstract}

\section{Introduction}

\begin{figure*}[ht!]
    \centering
    \includegraphics[width=1\linewidth]{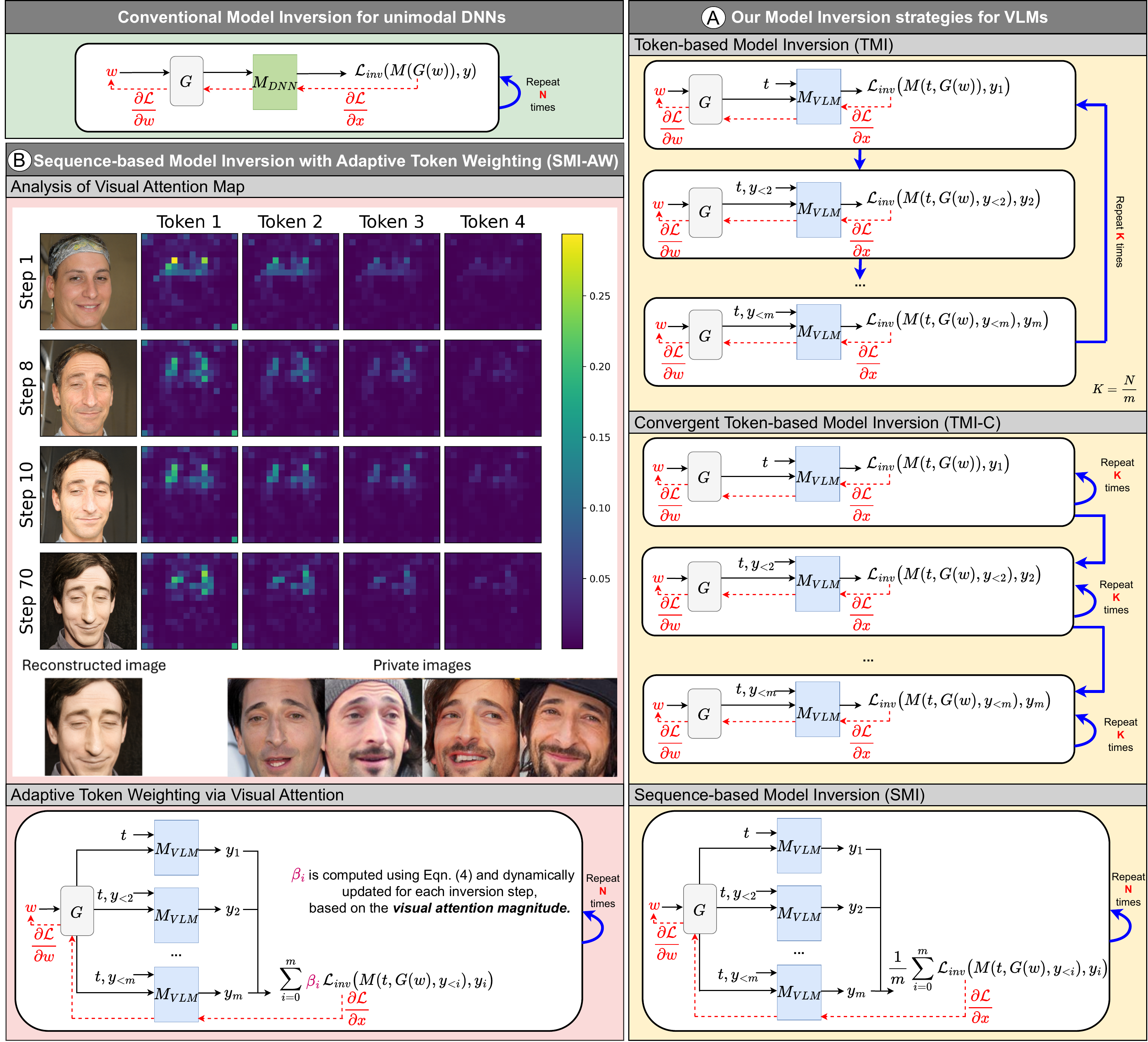}
    \caption{\textbf{We conduct the first systematic study of MI attacks on VLMs.}
{\bf (A)}
Designed for the token-generative characteristics of VLMs, we introduce 
a set of token-level and sequence-level MI strategies to investigate VLMs' privacy vulnerability 
(Sec.~\ref{sec:MI}).
Particularly, 
conventional MI typically targets unimodal DNNs, where the adversary seeks to reconstruct a training image $x = G(w)$ that maximizes the likelihood of a target class label $y$ under the target model $M_{DNN}$ by repeating $N$ inversion steps.
In contrast, VLMs $M_{VLM}$ generate a sequence of tokens, and the target output $\by = (y_1, \dots, y_m)$ is also a sequence of $m$ tokens. To address the unique nature of VLMs, we introduce several MI strategies:
Token-based Model Inversion (TMI), 
Convergent Token-based Model Inversion (TMI-C), and Sequence-based Model Inversion (SMI).
{\bf (B)}
Building on the insight that output tokens differ in their degree of visual grounding, and hence  their gradients vary in informativeness for reconstructing images during inversion, we propose Sequence-based Model Inversion with Adaptive Token Weighting (SMI-AW), a novel MI for VLMs (Sec.~\ref{sec:SMI-AW}).
SMI-AW adaptively adjusts each token's gradient contribution according to its visual grounding, allowing the optimization to concentrate on visually grounded tokens and more effectively recover private training images.
See Figure~\ref{fig:attentionMap} 
for discussion of attention map analysis.
}
\label{fig:proposed_method}
\end{figure*}

Model Inversion (MI) attacks aim to reconstruct training data by exploiting information encoded within a trained model. These attacks pose significant privacy risks to unimodal DNNs~\citep{fredrikson2015model, zhang2020secret, chen2021knowledge, an2022mirror, struppek2022plug, kahla2022label, han2023reinforcement,nguyen2023re, yuan2023pseudo,nguyen2023label,ho2024model,peng2024pseudo,qiu2024closer,tran2025random}, 
The goal of MI attack is to reconstruct private training images $x$ associated with a target label $y$. These methods typically pose inversion as an optimization problem that maximizes the likelihood of $y$ under the target model: 
\begin{equation}
\max_{w} \log \mathbb{P}_{M_{DNN}}(y\mid G(w))
\end{equation}
Here, $M_{DNN}$ is a unimodal DNN trained on private data $\mathcal{D}_{priv}$, and $G$ represents a generative model \citep{goodfellow2014generative,adler2018banach,karras2019style}. The optimization is usually accomplished by performing $N$ inversion update steps to generate a reconstruction $x^* = G(w^*)$ that approximates the training sample in $\mathcal{D}_{priv}$ for a given label $y$ (See Related Work in Supp).

\textbf{Research Gap.} 
With the rapid advancement and widespread deployment of Vision-Language Models across various applications \citep{zhu2023minigpt,liu2024llavanext,team2024gemma,yao2024minicpm,wang2025internvl3,hong2025glm}, an important and timely question arises: \emph{Are VLMs similarly vulnerable to Model Inversion attacks as unimodal DNNs?} In this context, we define an MI attack as the task of reconstructing VLM’s training images by leveraging its textual input and output. Addressing this question is crucial for understanding potential privacy threats in multimodal systems.

Unlike unimodal DNNs, vision-language models $M_{VLM}$ differ in several fundamental ways: they process multiple modalities (e.g., images and text), often comprise several distinct modules (e.g., separate encoders for vision and language, projector, language model), are often trained in multiple stages, and leverage broad, large-scale datasets. 
Crucially, a VLM's output is language, represented as a sequence of tokens.
Consequently, MI attacks on VLMs must contend with unique aspects not present in unimodal DNNs. 
Furthermore, in unimodal DNNs, private visual features are directly embedded in the model parameters, increasing the risk that model inversion attacks can extract private visual features directly from the model.
In contrast, many VLMs keep the vision encoder frozen during training and primarily update the language model. As a result, inversion attacks on VLMs rely on
private information embedded in 
the language model's and projector's parameters to guide the image reconstruction, rather than directly extracting visual features from the vision encoder.
These differences highlight a timely and important research gap: \textit{The urgent need for novel Model Inversion tailored to the multimodal VLMs to understand their privacy threats}.

\textbf{In this work}, 
we conduct the first systematic investigation of MI attacks on modern VLMs
(Figure~\ref{fig:proposed_method}).
The token-generative nature of VLMs necessitates new MI attack designs beyond conventional unimodal approaches. To this end, we introduce a suite of token-based and sequence-based inversion strategies.
Our token-based methods leverage token-level gradients to guide reconstruction, while our sequence-based methods aggregate gradients across the entire output sequence to provide a globally coherent optimization signal.
{\em Crucially}, this framework reveals a key insight: output tokens differ substantially in their degree of visual grounding, and thus in how informative their gradients are for reconstructing images. Building on this observation, we propose {\em Sequence-based Model Inversion with Adaptive Token Weighting} (SMI-AW), which dynamically reweights token contributions using their visual attention strength, producing visually relevant gradients and enabling more accurate reconstruction of private training images.


We conduct experiments on a range of VLMs across multiple datasets to demonstrate the effectiveness of our inversion attacks. Notably, human evaluation of the reconstructed images achieves an attack accuracy of 61.21\%, highlighting the severity of model inversion threats in VLMs. Furthermore, we validate the generalizability of our approach on publicly available VLMs, reinforcing its practical applicability and security implications. 
Our key contributions are as follows:
\begin{itemize}
\item We present a pioneering study of MI attacks on VLMs, uncovering a  security risk in the multimodal models.

\item We introduce a suite of inversion strategies tailored for
token-generative nature of VLMs (Sec.~\ref{sec:MI}).

\item Based on our observation that output tokens' gradients differ in their informativeness for MI, we propose SMI-AW, which dynamically reweights token contributions in different inversion steps (Sec.~\ref{sec:SMI-AW}).

\item The extensive experimental validation shows our proposed attacks, especially SMI-AW, achieve both good attack accuracy and good visual fidelity. Crucially, we showcase successful inversion attacks against publicly available VLMs, underscoring the immediate and practical privacy risks posed by these models (Sec.~\ref{sec:experiment}).
\end{itemize}

\section{Problem Formulation}



\textbf{Threat Model.}
We consider a threat model where a VLM $M$
is trained 
on a private VQA dataset $\mathcal{D}_{priv} = \{(t, \bx, y)\}$, where $\bx$ is the image, $t$ and $y$ are the textual input and correct textual answer.
For clarity, hereafter we use $M$ to denote a VLM and $M_{DNN}$ to refer to a unimodal DNNs. 
Using the tokenizer of $M$, the textual input $t$ and the textual answer $y$ are tokenized into sequences $\bt = (t_1, t_2, \dots, t_n)$ and $\by = (y_1, y_2, \dots, y_m)$, respectively.
We denote the full output sequence of $M$ given input $(\bt, \bx)$ as $M(\bt, \bx)$. The model’s prediction of the $i$-th token $y_i$, conditioned on the previous tokens $y_{<i}$, is denoted by $M(\bt, \bx, y_{<i})$.

\textbf{Attacker's Goal.}
Given a trained VLM $M$, 
the goal of a model inversion attack is to reconstruct a representative image $\bx^*$ that reveals sensitive or private visual information from the private training image $\bx$ in a data sample $(t, \bx, y) \in \mathcal{D}_{priv}$. 
Specifically, the adversary is given access to the trained model $M$, a textual input prompt $t$, and the corresponding target output $y$.
The attacker's goal is to infer a plausible visual input $\bx^*$ that
produces the high likelihood output sequence $\by$ under the  input tokens $\bt$.
This reconstructed image $\bx^*$ is intended to approximate or reveal private features of the true image $\bx$, thereby compromising the visual confidentiality of the training data.

\textbf{Attacker's Capabilities.}
We consider a white-box setting \citep{zhang2020secret,chen2021knowledge,an2022mirror,struppek2022plug,nguyen2023re,qiu2024closer}, where the attacker has full access to the VLM's architecture, parameters, attention maps, output responses (e.g., generated text or logits).




\section{Model Inversion Strategies for VLMs}
\label{sec:MI}


We consider a  VLM $M$ trained on 
a private VQA dataset 
$\mathcal{D}_{priv} = \{(t,\bx,y)\}$.  
Performing MI attacks directly in the image space is computationally expensive and often ineffective \citep{zhang2020secret}. To reduce the search space of $x^*$, we follow conventional MI approaches for DNNs by leveraging a generative model $G$ trained on an auxiliary public dataset 
$\mathcal{D}_{pub}$ \citep{zhang2020secret,chen2021knowledge,struppek2022plug,nguyen2023re,qiu2024closer}. This allows us to shift the optimization from the high-dimensional image space to the lower-dimensional latent space of $G$, i.e., $x = G(w)$, where $w$ is the intermediate latent vector.

In contrast to conventional MI attacks targeting classification models, where the objective is to reconstruct an input image $x$ that yields a specific class label, \textit{VLMs generate token sequences, and the target output also represented as a sequence of tokens}. This requires a reformulation of the MI objective to account for token generation. 
In this section, we introduce new token-based and sequence-based MI methods. In
Sec.~\ref{sec:SMI-AW}, we further propose a novel MI with dynamic weighting to account for varying informativeness of different tokens' gradients during inversion.

 

\subsection{Token-based Model Inversion (TMI)}

\begin{algorithm}[t!]
            \caption{\textbf{Token-based MI (TMI)}}
            \begin{algorithmic}[1]
            \State \textsc{\textbf{Input:}} $M, G, \bt,\by=(y_1,\dots,y_m),N,\lambda$
            \State \textsc{\textbf{Output:}} $G(w)$
            \State $K=N/m$
            \For{$k = 1$ to $K$}
                \For{$i = 1$ to $m$}   
                                    
                    \State \begin{minipage}{0.8\linewidth} 
                    \vspace{-0.2cm}
                    \begin{equation}
                    \hspace{-1.6cm}
                    \mathcal{L} = \mathcal{L}_{inv}(M(\bt,G(w),y_{<i}),y_i)
                    \label{eq:L_TMI}
                    \end{equation}  
                    \vspace{-0.4cm}
                    \end{minipage}
            
                    \State $w = w - \lambda\frac{\partial\mathcal{L}}{\partial w} $
                \EndFor
            \EndFor
            \end{algorithmic}
            \label{alg:TMI}
        \end{algorithm}   

A natural approach is to treat the inversion process as a sequential update over individual token predictions. Given a target token sequence $\by$, we iteratively update the latent code $w$ after each generated token (see Figure \ref{fig:proposed_method} (A) TMI). The details are in Algorithm \ref{alg:TMI}.  $N$ is the number of inversion steps, $\lambda$ is the update rate of MI, $y_{<i}$ denotes the previous tokens. $\mathcal{L}_{inv}$ presents the inversion loss, guiding the generative model $G$ to produce images that induce the token $y_i$. We discuss the design of $\mathcal{L}_{inv}$ in the Supp.
The optimization is performed over multiple iterations, typically up to a  update limit of $N$ inversion steps. At each iteration, each token contributes independent update to $w$.



\subsection{Convergent Token-based Model Inversion (TMI-C)}

TMI performs a single update per token per iteration. However, VLMs generate each token $y_i$ based on the preceding tokens $y_{<i}$.
To better align with this generative dependency, we propose Convergent Token-based Model Inversion (TMI-C), which updates the latent vector $w$ multiple times for each target token before proceeding to the next. Specifically, for each token $y_i$, we perform $K$ updates to $w$, thereby encouraging convergence of the token-level inversion subproblem before advancing to $y_{i+1}$  (see Figure \ref{fig:proposed_method} (A) TMI-C). The details are presented in Algorithm~\ref{alg:TMIC}.

    \begin{algorithm}[ht!]
        \caption{\textbf{Convergent Token-based MI (TMI-C)}}
            \begin{algorithmic}[1]
            \State \textsc{\textbf{Input:}} $M, G, \bt,\by=(y_1,\dots,y_m),N,\lambda$
            \State \textsc{\textbf{Output:}} $G(w)$
            \State $K=N/m$
            \For{$i = 1$ to $m$}  
                \For{$k = 1$ to $K$}
                    \State Compute $\mathcal{L}$ using Eqn. (\ref{eq:L_TMI}). 
                    \State $w = w - \lambda\frac{\partial\mathcal{L}}{\partial w} $
                \EndFor
            \EndFor
            \end{algorithmic}
            \label{alg:TMIC}  
    \end{algorithm}

\subsection{Sequence-based Model Inversion (SMI)}

Token-based model inversion methods treat each token independently, optimizing the latent vector $w$ based on individual token-level losses. 
As the output of VLMs is a sequence of tokens, 
we propose Sequence-based Model Inversion (SMI), which performs a single gradient update to $w$ by averaging the loss across all $m$ tokens in the sequence  (see Figure \ref{fig:proposed_method} (A) SMI). By aggregating token-level losses into a unified objective, SMI leverages the interdependencies among tokens and provides more coherent gradients that reflects the structure of the full sequence. This global view encourages the model to  recover a latent representation that is consistent across the entire sequence, rather than optimizing for each token in isolation. The details are presented in Algorithm~\ref{alg:SMI}.

     \begin{algorithm}[ht!]
    \caption{\textbf{Sequence-based MI (SMI)}}
    \begin{algorithmic}[1]
    \State \textsc{\textbf{Input:}} $M, G, \bt,\by=(y_1,\dots,y_m),N,\lambda$
    \State \textsc{\textbf{Output:}} $G(w)$
    \For{$k = 1$ to $N$}  
        
    \State    
    \hspace{-0.3cm}
    \begin{minipage}{0.9\linewidth}   
        \begin{equation}
        \hspace{-1cm}
        \mathcal{L} = \frac{1}{m} \sum_{i=1}^m \mathcal{L}_{inv}(M(\bt, G(w), y_{<i}), y_i)
        \label{eq:L_SMI}
        \end{equation}     
        \end{minipage}
        
    \State $w = w - \lambda\frac{\partial\mathcal{L}}{\partial w} $
    \EndFor
    \end{algorithmic}
    \label{alg:SMI}
\end{algorithm}

\section{Sequence-based Model Inversion with Adaptive Token Weighting (SMI-AW)}
\label{sec:SMI-AW}

\begin{figure}[ht!]
    \centering
\includegraphics[width=0.99\linewidth]{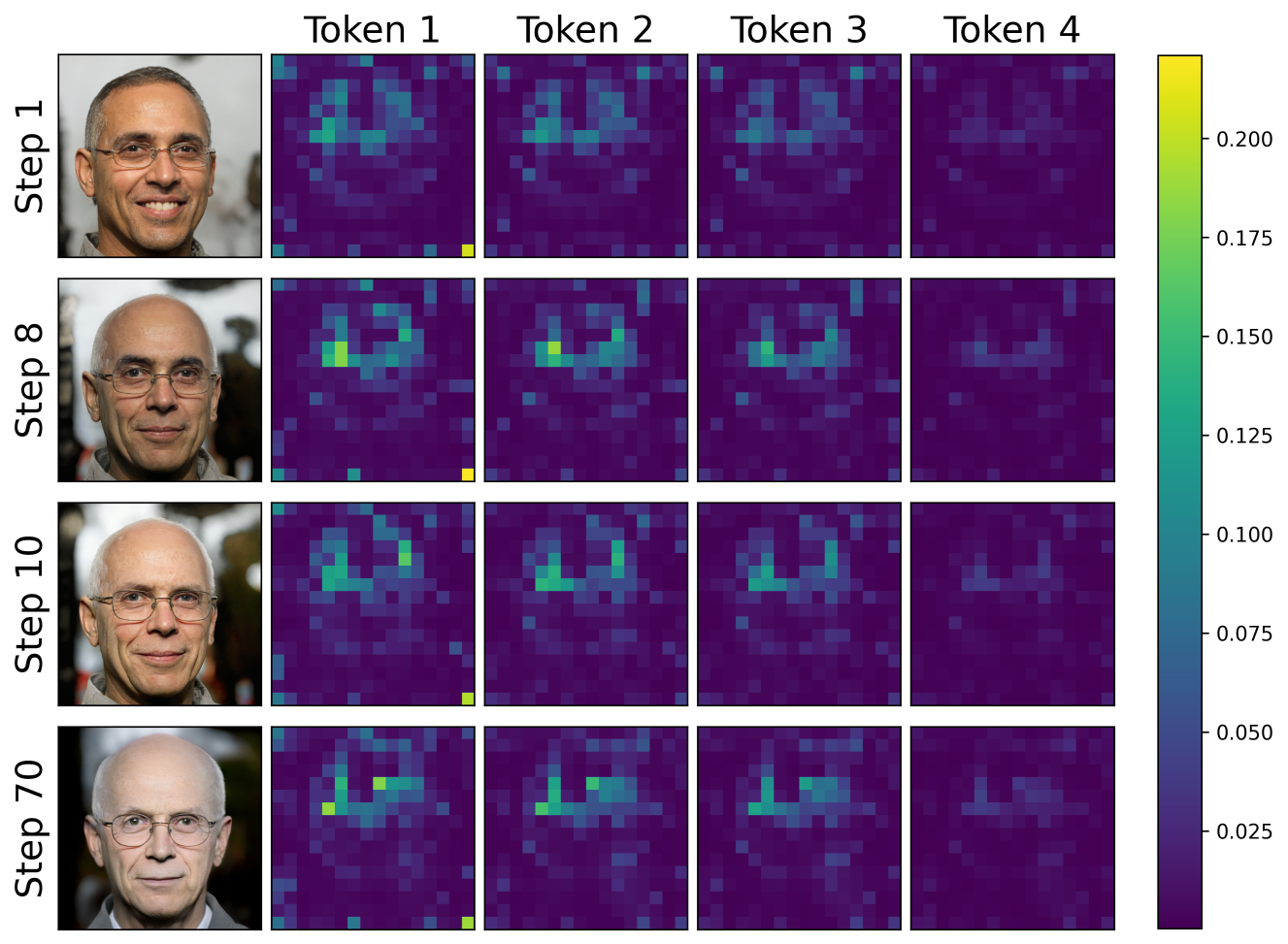}
    \includegraphics[width=0.87\linewidth]{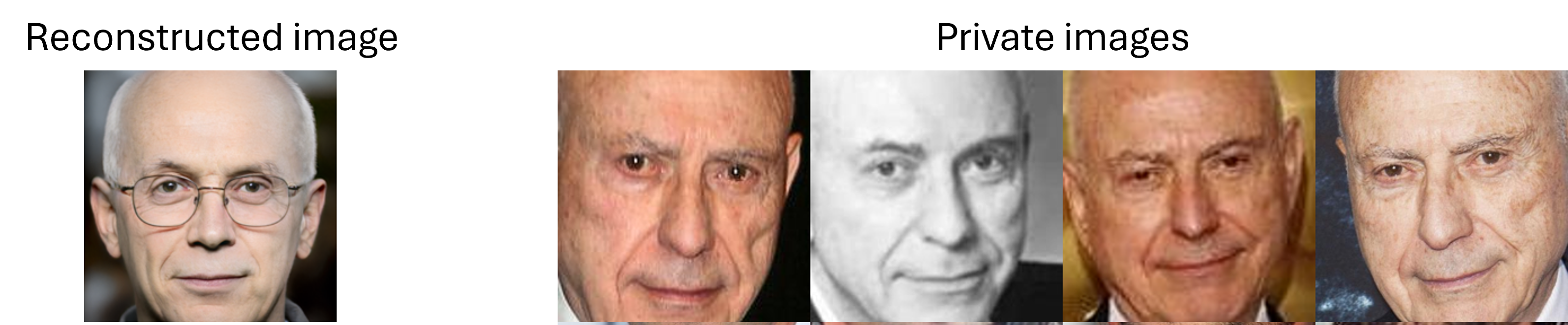}    
\rule{1.0\linewidth}{0.4pt}\\    
 \includegraphics[width=0.99\linewidth]{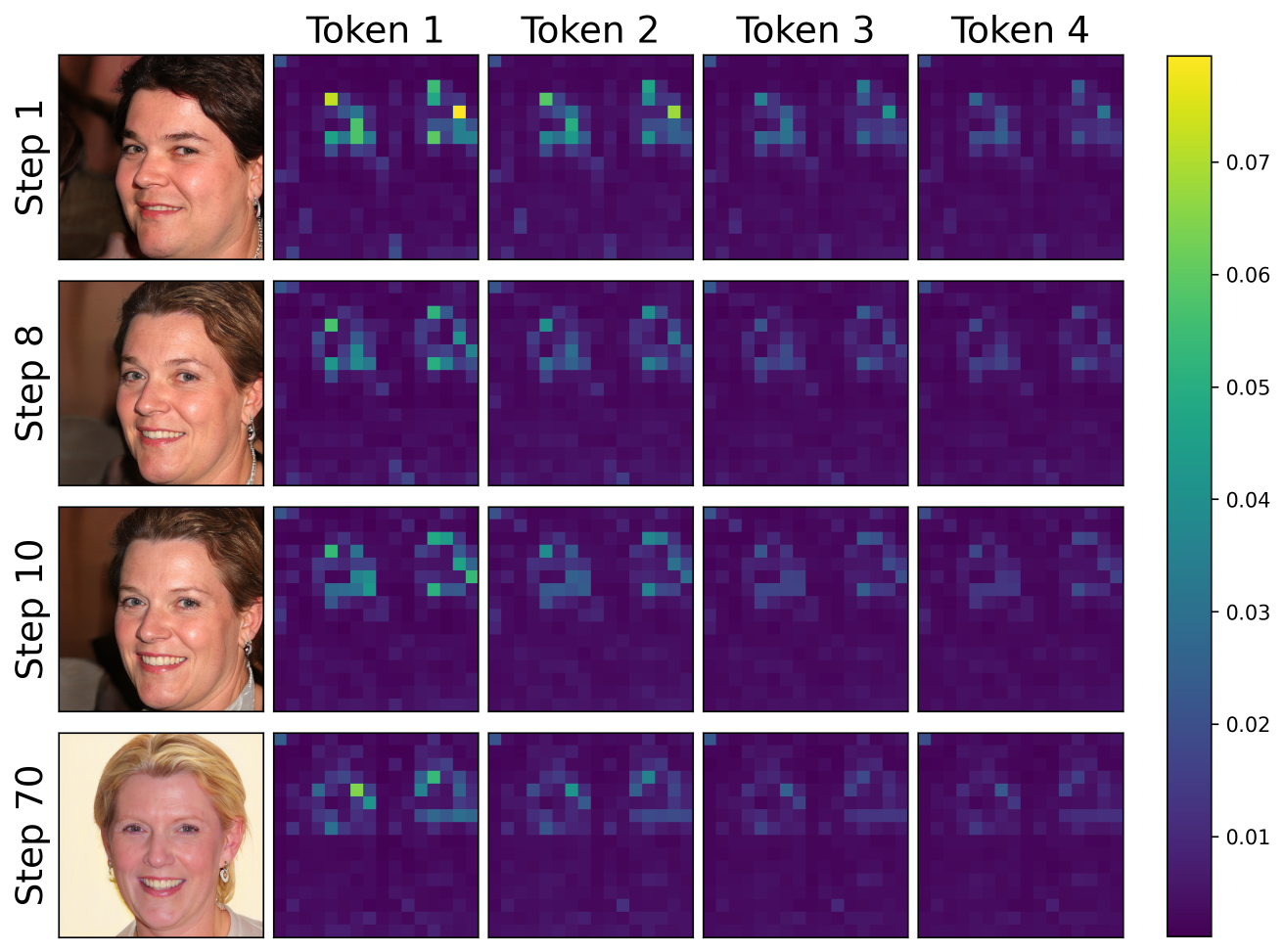}
    \includegraphics[width=0.87\linewidth]{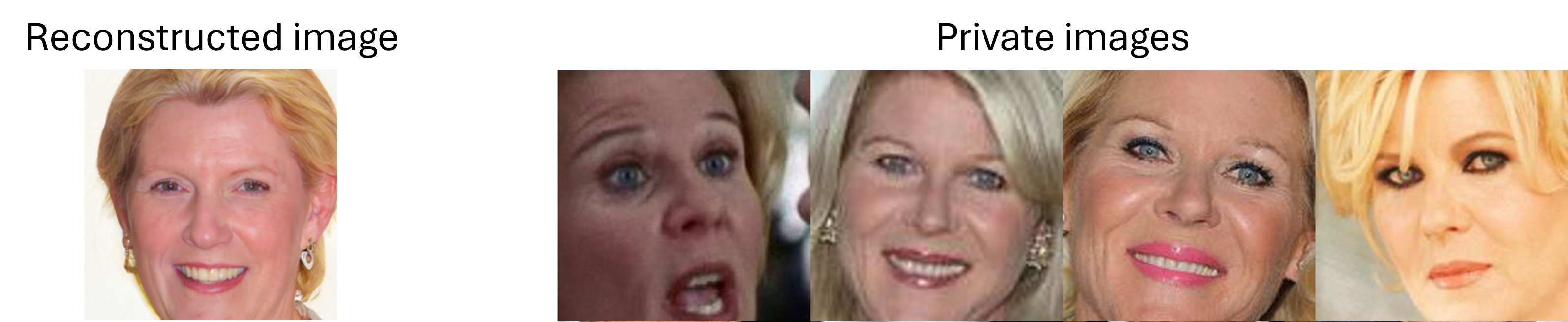}
    \caption{
{\bf Analysis of visual–textual attention across output tokens and inversion steps.}
We visualize the cross-attention map between the reconstructed image and each output token during inversion. Different tokens exhibit markedly different attention maps: visually grounded tokens show strong attention, while others produce weak responses, indicating limited reliance on the image. Moreover, attention patterns evolve over inversion steps, as a token's dependence on visual input changes when the reconstructed image becomes more consistent with the target output. These observations reveal that token-level gradients vary substantially in visual informativeness both across tokens and over time. This motivates our SMI-AW method, which dynamically reweights token contributions based on their visual attention strength.
{\bf Additional attention map analysis can be found in Supp.}
    }
    \label{fig:attentionMap}
\end{figure}

In this section, 
we further propose 
a novel sequence-based MI with dynamic weighting.
VLMs generate each output token $y_i$ based on the preceding text tokens $y_{<i}$
and the image
$\bx$. 
We observe that different output tokens have varying degrees of dependence on the visual input — some are strongly visually grounded, while others are less
visually grounded and they are driven by prior linguistic context instead (Figure
\ref{fig:proposed_method} (B), Figure
\ref{fig:attentionMap}).
Consequently, 
{\em the gradients of 
 output token $y_i$ 
vary in informativeness for reconstructing images during MI.}

If a token
$y_i$
exhibits strong visual attention, it is likely more visually dependent, and its loss gradient carries richer visual information about the image. In other words, the strength of cross-attention 
could indicate 
how sensitive the token’s prediction is to the image content, which directly determines how informative its gradient is for model inversion. Therefore, we propose to use the magnitude of the attention map as a proxy for the informativeness of each token's loss gradient in a model inversion step and use it to weight its contribution to the overall inversion gradient — tokens with higher visual attention receive larger weights, while those with weaker visual grounding are down-weighted.
Note that the magnitude of the attention map can be readily obtained in white-box MI.

Let
$\alpha_i$
denote the total visual attention value of the output token $y_i$. 
The corresponding weight $\beta_i$ for each output token $y_i$ is then computed as:

\begin{equation}
\beta_i = \frac{\alpha_i}{\sum_{j=1}^m \alpha_j}
\label{eq:alpha}
\end{equation}
Furthermore, we update these weights $\beta_i$ dynamically across inversion steps, since a token’s dependence on visual input can change as the reconstructed image gradually becomes more consistent with the target output. 
The method is presented in Algorithm \ref{alg:SMIAW}. 
{\em Overall, this adaptive weighting enables the optimization 
to focus 
on visually-grounded output tokens, producing gradients that more effectively guide the reconstruction of the private training image.}

\noindent
    \begin{algorithm}[ht!]
    \caption{\textbf{Sequence-based MI with Adaptive Token Weighting (SMI-AW)}}
    \begin{algorithmic}[1] 
    \State \textsc{\textbf{Input:}} $M, G, \bt, \by=(y_1,\dots,y_m), N, \lambda$
    \State \textsc{\textbf{Output:}} $G(w)$
    \For{$k = 1$ to $N$}  
        \State Compute $\beta_i$ for each token $y_i$ using Eqn. (\ref{eq:alpha})

    \State    \hspace{-0.6cm}\begin{minipage}{0.9\linewidth}    
        \vspace{-0.2cm}
        \begin{equation}
        \mathcal{L} = \sum_{i=1}^m \beta_i\mathcal{L}_{inv}(M(\bt, G(w), y_{<i}), y_i)
        \label{eq:L_SMIAW}
        \end{equation}     
        \end{minipage}
        
            \State $w = w - \lambda\frac{\partial\mathcal{L}}{\partial w} $
    \EndFor
    \end{algorithmic}
    \label{alg:SMIAW}
    \end{algorithm}


\begin{tcolorbox}[
    colback=white,      
    colframe=black,     
    boxrule=0.5pt,      
    arc=2mm,            
    left=6pt, right=6pt, top=4pt, bottom=4pt, 
]
\textbf{Remark.} To tailored for VLMs’ token-based generative nature, we propose four token-based and sequence-based that leverage token-level and sequence-level gradients for image reconstruction. Notably, SMI-AW, our novel MI method for VLMs, focuses the optimization on visually grounded tokens. 
\end{tcolorbox}
\section{Experiments}
\label{sec:experiment}
In this section, we evaluate the effectiveness of our 4 proposed MI attacks on 4 VLMs (i.e., LLaVA-v1.6\citep{liu2024llavanext}, Qwen2.5-VL\citep{Qwen2.5-VL}, MiniGPT-v2\citep{chen2023minigptv2}, InternVL2.5\citep{chen2024internvl,chen2024expanding}), 3 private datasets, 2 public datasets with an extensive evaluation spanning 5 metrics including the human evaluation.

\subsection{Experimental Setting}\label{Experimental_Setting}

\paragraph{Dataset.}
Following standard model inversion (MI) setups\citep{zhang2020secret,chen2021knowledge,struppek2022plug,an2022mirror,nguyen2023re,yuan2023pseudo,struppekcareful,qiu2024closer,ho2024model,koh2024vulnerability}, we use facial and fine-grained classification datasets to evaluate our approach. Specifically, we conduct experiments on three datasets: FaceScrub\citep{ng2014data}, CelebA\citep{liu2015deep}, and StanfordDogs\citep{dataset2011novel}. FaceScrub dataset contains 106,836 images across 530 identities. For CelebA, we select the top 1,000 identities with the most samples from the full set of 10,177 identities. StanfordDogs comprises images from 120 dog breeds, serving as a representative fine-grained visual dataset.

To train the target VLMs, we construct VQA-style datasets including VQA-FaceScrub, VQA-CelebA, and VQA-StanfordDogs. For the facial datasets, each image $x$ is paired with a prompt $t =$ \textit{``Who is the person in the image?''}, and the expected textual response $y$ is the individual's name (e.g., $y =$ ``Candace Cameron Bure"). Since the CelebA dataset does not contain identity names, we randomly generate 1,000 unique English names, each comprising a distinct first and last name with no repetitions, and assign one to each identity in the selected CelebA subset.
For VQA-StanfordDogs, each image $\bx$ is paired with a prompt $t =$ \textit{``What breed is this dog?''}, and the target answer $y$ corresponds to the ground-truth breed label (e.g., ``black-and-tan coonhound").

\paragraph{Public Dataset and Image Generator.}
For facial image reconstruction, we use FFHQ\citep{karras2019style} as the public dataset $\mathcal{D}_{pub}$ and a pre-trained StyleGAN2\citep{karras2020analyzing} trained on FFHQ. Following conventional MI~\citep{struppek2022plug}, we optimize in the latent space $w$ of StyleGAN2 to recover images $x = G(w)$. For StanfordDogs experiments, we adopt AFHQ-Dogs\citep{choi2020stargan} as $\mathcal{D}_{pub}$ to train the dog image generator.

\paragraph{VLMs.}
We fine-tune LLaVA-v1.6-7B\citep{liu2024llavanext}, Qwen2.5VL-7B\citep{Qwen2.5-VL}, MiniGPT-v2\citep{chen2023minigptv2}, and InternVL2.5-8B\citep{chen2024internvl,chen2024expanding} using VQA-Facescrub, VQA-CelebA, and VQA-StanfordDogs. These models are selected to cover a diverse spectrum of architectures, projection designs, and training paradigms.

\paragraph{Inversion Loss Design for VLMs.}
We extend the inversion loss from conventional unimodal MI to VLMs. 
Specifically, we adopt three widely used identity losses in traditional MI to MI for VLMs: the cross-entropy loss $\mathcal{L}_{CE}$\citep{zhang2020secret,chen2021knowledge,qiu2024closer}, the max-margin loss $\mathcal{L}_{MML}$\citep{yuan2023pseudo}, and the logit-maximization loss $\mathcal{L}_{LOM}$\citep{nguyen2023re}. Detailed formulations are provided in the Supp.

\paragraph{Evaluation Metrics.}
To assess the quality of the inversion results, we adopt five metrics:
\begin{itemize}
    \item \textbf{Attack accuracy.} We compute the attack accuracy using three frameworks as described below. We strictly follow the evaluation frameworks in their original works ( detailed setups in the Supp). Higher accuracy indicates a more effective inversion attack.
    \begin{itemize}
        \item \textbf{Attack accuracy evaluated by conventional evaluation framework $\mathcal{F}_{DNN}$ ($AttAcc_{D} \uparrow$)}\citep{zhang2020secret,chen2021knowledge,struppek2022plug,nguyen2023re,qiu2024closer}. This is a conventional framework, where the evaluation models are standard DNNs trained on private dataset. Following\citep{struppek2022plug,struppekcareful}, we use InceptionNet-v3\citep{szegedy2016rethinking} as the evaluation model to classify reconstructed images, and compute the $Top 1$ and $Top 5$ based on whether the predicted label match the target label.         
        \item  \textbf{Attack accuracy evaluated by MLLM-based evaluation framework $\mathcal{F}_{MLLM}$ ($AttAcc_{M} \uparrow$)}.\citep{ho2025revisitingmodelinversionevaluation} demonstrate that $\mathcal{F}_{MLLM}$ can achieve better alignment with human evaluation. 
        Unlike the conventional framework $\mathcal{F}_{DNN}$, which relies on the classification predictions of standard DNNs trained on private datasets, this metric leverages powerful MLLMs to evaluate the success of MI-reconstructed by referencing the corresponding private images.        
        \item \textbf{Attack accuracy evaluated by human $\mathcal{F}_{Human} (AttAcc_{H} \uparrow)$}. Following existing studies\citep{an2022mirror,nguyen2023re}, we conduct the user study on Amazon Mechanical Turk. Participants are asked to evaluate the success of MI-reconstructed by referencing the corresponding private images (Details in the Supp).
    \end{itemize}
    \item \textbf{Feature distance.} We compute the $l_2$ distance between the feature representations of the reconstructed and the private training images\citep{struppek2022plug}. Lower values indicate higher similarity and better inversion quality.
    \begin{itemize}
        \item \textbf{$\delta_{eval}$.} Features are extracted by the evaluation model in $\mathcal{F}_{DNN}$.
        \item \textbf{$\delta_{face}$.} Features are extracted by a pre-trained FaceNet model\citep{schroff2015facenet}.
    \end{itemize} 
\end{itemize} 

\subsection{Results}
\label{sec:results}

We report attack results on the FaceScrub dataset in Table~\ref{tab:Facescrub_Results}, evaluating four MI strategies under three inversion losses using LLaVa-1.6-7B.
The results show that sequence-based mode inversion methods consistently outperform token-level MI approaches across all evaluation metrics. Among them, SMI-AW, when combined with the $\mathcal{L}_{LOM}$, achieves the highest performance. This highlights the advantage of employing adaptive token-wise weights that are dynamically updated at each inversion step. 
Using this method, we achieve an attack accuracy of 61.01\% under $\mathcal{F}_{MLLM}$ while other distance metrics such as $\delta{face}$ and $\delta_{eval}$ are the lowest (where lower is better).

Results on additional datasets, including CelebA and StanfordDogs, are shown in Table~\ref{tab:celeba_dogs} using the logit maximization loss. We achieve high attack success rates, with attack accuracies of 67.05\% on CelebA and 78.13\% on StanfordDogs. These findings are consistent with results on the FaceScrub dataset, where SMI-AW consistently achieves the highest attack performance across all metrics.

We further evaluate our proposed method SMI-AW on
Qwen2.5VL-7B\citep{Qwen2.5-VL}, MiniGPT-v2\citep{chen2023minigptv2}, and InternVL2.5-8B\citep{chen2024internvl,chen2024expanding} 
(see Table \ref{tab:Facescrub_QWEN2.5VL_Results}). The results reinforce the generalizability of our findings, demonstrating that VLMs are broadly vulnerable to model inversion attacks. These results underscore the severity of this vulnerability and raise a significant alarm about the susceptibility of VLMs to inversion-based privacy breaches.

\begin{table}[]
    \centering
\caption{Comparison of performance metrics across four inversion strategies using LLaVa-1.6-7B fine-tuned on the FaceScrub dataset, evaluated with three identity losses. We highlight the best results in bold.
}
\begin{adjustbox}{width=1.0\columnwidth}
\begin{tabular}{lcc rcc}
\hline
\multirow{2}{*}{$\mathcal{L}_{inv}$} & \multirow{2}{*}{$AttAcc_{M}\uparrow$} & \multicolumn{2}{c}{$AttAcc_{D} \uparrow$} &  \multirow{2}{*}{$\delta_{face}\downarrow$} &  \multirow{2}{*}{$\delta_{eval}\downarrow$} \\ \cline{3-4}
 &  & \multicolumn{1}{c}{$Top 1$} & \multicolumn{1}{c}{$Top 5$} &  &   \\ \hline

\multicolumn{6}{c}{\textbf{TMI}}  \\ \hline
$\mathcal{L}_{CE}$ & 37.78\% & 17.71\% & 39.79\% & 0.8939 & 147.35 \\
$\mathcal{L}_{MML}$ & 39.98\% & 17.31\% & 38.51\% & 0.9065 & 193.14 \\
$\mathcal{L}_{LOM}$ & 44.34\% & 21.77\% & 44.69\% & 0.8488 & 141.87 \\ \hline

\multicolumn{6}{c}{\textbf{TMI-C}}  \\ \hline
$\mathcal{L}_{CE}$ & 21.77\% & 6.39\% & 18.58\% & 1.0911 & 636.50 \\
$\mathcal{L}_{MML}$ & 25.99\% & 6.51\% & 18.82\% & 1.0659 & 205.71 \\
$\mathcal{L}_{LOM}$ & 31.16\% & 9.32\% & 24.22\% & 1.0221 & 457.49 \\ \hline
 
\multicolumn{6}{c}{\textbf{SMI}}  \\ \hline
$\mathcal{L}_{CE}$ & 40.97\% & 18.25\% & 41.11\% & 0.8682 & 144.53 \\
$\mathcal{L}_{MML}$ & 55.52\% & 32.83\% & 60.12\% & 0.7569 & 137.43 \\
$\mathcal{L}_{LOM}$ & 59.17\% & 33.47\% & 61.89\% & 0.7465 & 140.83 \\ \hline
 

\multicolumn{6}{c}{\textbf{SMI-AW}}  \\ \hline
$\mathcal{L}_{CE}$ & 41.16\%  & 18.71\%	& 43.04\%	& 0.8782 &	143.95  \\
$\mathcal{L}_{MML}$ & 56.23\% & 35.83\%	&	62.50\% &	0.7451	& 138.03\\
$\mathcal{L}_{LOM}$ & \textbf{
61.01}\% & \textbf{37.62}\%	& \textbf{66.16}\%	&\textbf{ 0.7265}	& \textbf{134.94}  \\ \hline
\end{tabular}
\end{adjustbox}
\label{tab:Facescrub_Results}

\end{table}

\begin{table}[]
    \centering
    \caption{We report the results on the CelebA and StanfordDogs dataset across four inversion strategies with $\mathcal{L}_{LOM}$.}

    \begin{adjustbox}{width=1.0\columnwidth}
    \begin{tabular}{lllccc}
    \hline        
    
    \multirow{2}{*}{\textbf{Method}} & \multirow{2}{*}{$AttAcc_{M}\uparrow$} & \multicolumn{2}{c}{$AttAcc_{D}\uparrow$} &  \multirow{2}{*}{$\delta_{face}\downarrow$} &  \multirow{2}{*}{$\delta_{eval}\downarrow$} \\ \cline{3-4}
      &  & \multicolumn{1}{c}{$Top 1$} & \multicolumn{1}{c}{$Top 5$} &  &   \\ \hline          
    \multicolumn{6}{c}{\textbf{CelebA dataset}} \\ \hline        
    TMI & 39.74\% & 15.31\% & 33.14\% & 1.0195 & 428.66 \\
    TMI-C & 18.73\% & 3.63\% & 10.29\% & 1.2370 & 446.90 \\ 
    SMI & 64.93\% & 38.30\% & 63.69\% & 0.8294 & 416.34 \\
    SMI-AW & \textbf{67.05}\%  & \textbf{45.25}\%	 & \textbf{69.55}\% &	\textbf{0.8001} &	\textbf{413.90} \\\hline
    \multicolumn{6}{c}{\textbf{StanfordDogs dataset}} \\ \hline     
    TMI & 61.46\% & 40.31\% & 70.21\% & - & 102.40 \\
    TMI-C & 48.54\% & 29.69\% & 59.79\% & - &  102.23 \\
    SMI & 75.94\% & 53.65\% & 82.19\% & - & \textbf{76.98} \\
    SMI-AW & \textbf{78.13}\% & \textbf{56.15}\% & \textbf{84.79}\%	& - & 81.66  \\\hline
    \end{tabular}
    \label{tab:celeba_dogs}
    \end{adjustbox}
\end{table}

\begin{table}[]
    \centering
        
    \caption{We report the results of Qwen2.5-VL-7B, MiniGPT-v2, and InternVL2.5-8B on the Facescub dataset. Here we use SMI-AW with $\mathcal{L}_{LOM}$. }
    \begin{adjustbox}{width=1.0\columnwidth}
    \begin{tabular}{lccccc}
    \hline
    \multirow{2}{*}{$M$} &  \multirow{2}{*}{$AttAcc_{M}\uparrow$} & \multicolumn{2}{c}{$AttAcc_{D}\uparrow$} &  \multirow{2}{*}{$\delta_{face}\downarrow$} &  \multirow{2}{*}{$\delta_{eval}\downarrow$} \\ \cline{3-4}
      &  &  \multicolumn{1}{c}{$Top 1$} & \multicolumn{1}{c}{$Top 5$} &  &   \\ \hline
      

    \multicolumn{1}{l}{MiniGPTv2} &  47.92\% & 14.62\% &	33.82\%	& 0.9043	& 161.25\\
    Qwen2.5-7B & 32.03\%  & 13.21\%	& 27.24\% &  1.1308 & 150.46 \\
    InternVL2.5-8B & 55.05\% &	25.05\% &	52.10\% &	0.9185	& 139.18\\
    \hline
    \end{tabular}
    \end{adjustbox}
    \label{tab:Facescrub_QWEN2.5VL_Results}
\end{table}

\subsection{Analysis: Token-based vs. Sequence-based MI}

Our results show that token-based MI methods consistently underperform compared to sequence-based methods. There are two main reasons:
\begin{itemize}
\item
First, in token-based MI, gradients computed from a single output token can exhibit high variance and  be dominated by local linguistic context, making them noisy and unstable; consequently, an inversion step may be driven by an unstable signal that can misguide the optimization. 
\item
Second, some output tokens are only weakly visually grounded, as shown in our analysis in
Fig.~\ref{fig:attentionMap}.
Therefore, their gradients contain little information about the underlying image\cite{chen2024image,yang2025mitigating,chen2025spatial}. Updating the latent code based on such weakly informative tokens could lead to inconsistent or contradictory gradient directions across the sequence. 
\end{itemize}

Sequence-based MI (SMI) mitigates these issues by aggregating losses over the entire output sequence, producing a more stable and semantically coherent gradient direction that better reflects the visual content. However, SMI treats all tokens as equally informative, 
which is suboptimal because tokens differ substantially in their degree of visual grounding.
Our SMI-AW method further improves upon SMI by dynamically reweighting token contributions according to their visual attention strength, amplifying gradients from visually grounded tokens while suppressing noise from linguistically driven ones, achieving more  effective inversion updates.

To further analyze the difference between token-based and sequence-based MI, 
we examine the {\em match rate} between the final reconstructed images $M(\bt, G(w^*))$ and the corresponding target textual answers $y$. Specifically, we define the match rate as the percentage of reconstructed images for which the target answer $y$ appears as a substring of the predicted text associated with the image. In other words, it reflects the proportion of reconstructions whose generated text aligns with the target textual answer at the end of the inversion process.

\begin{figure}[t!]
    \centering
    \includegraphics[width=0.95\linewidth]{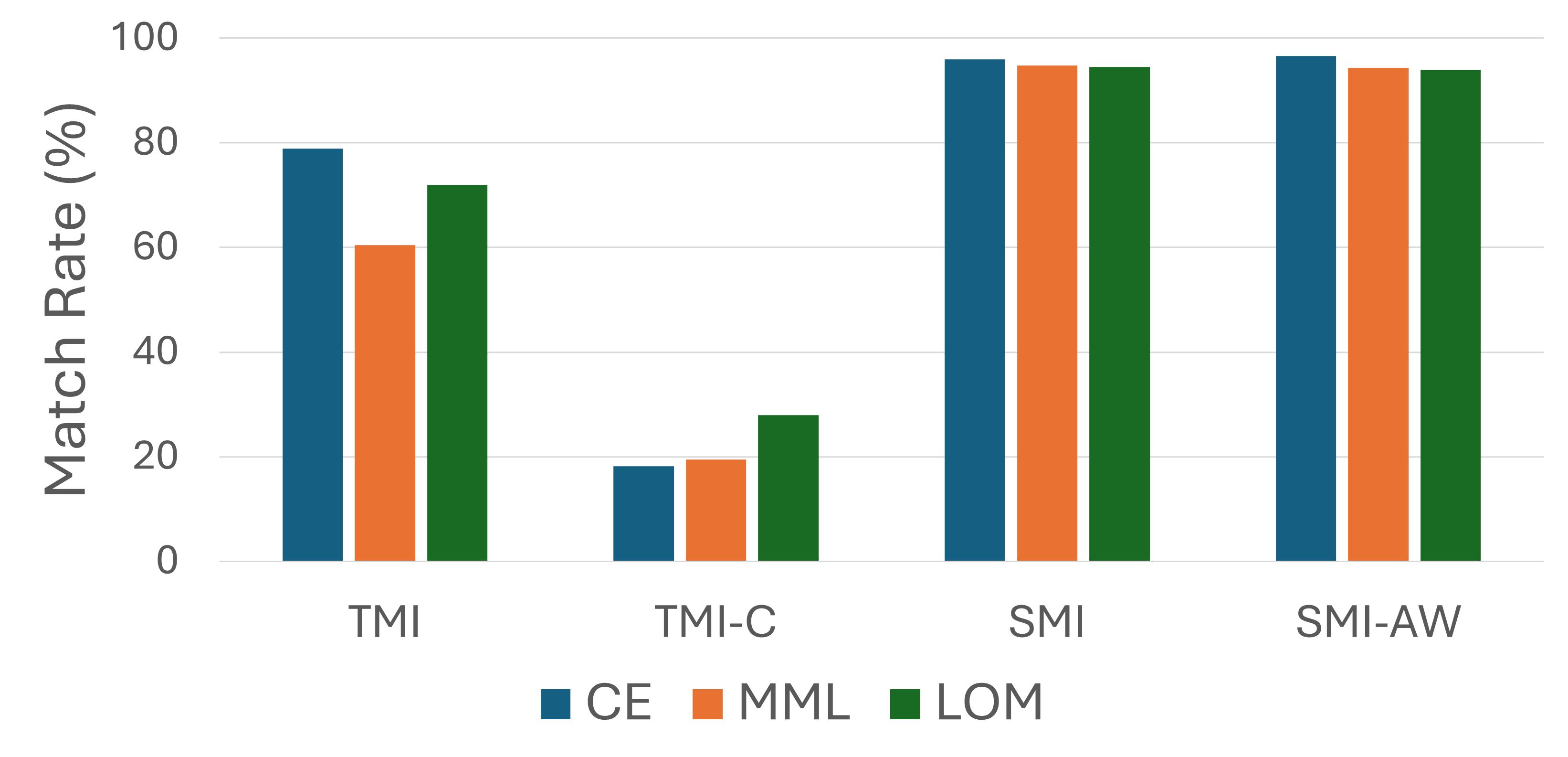}
    \caption{The match rate between the output text of the reconstructed image and the target output text $y$. 
}
\label{fig:match_rate}
\end{figure}

The results, shown in Figure~\ref{fig:match_rate}, reveal a clear distinction between the two types of methods. Token-based MIs, which perform inversion update with potentially unstable and weakly informative signals, 
exhibit poor convergence behavior, with match rates ranging from 60\% to 79\% for TMI, and dropping below 30\% for TMI-C. In contrast, sequence-based methods such as SMI and SMI-AW achieve match rates exceeding 95\%, indicating more reliable alignment between reconstructed images and their textual targets. 
It is important to note that a high match rate does not necessarily imply a successful attack, as the optimization may overfit or converge to a poor local minimum. Nevertheless, a higher match rate generally correlates with a greater likelihood of a successful identity inversion attack.

\subsection{Qualitative Results}

\begin{figure*}[t!]
    \centering
    \includegraphics[width=1.0\linewidth]{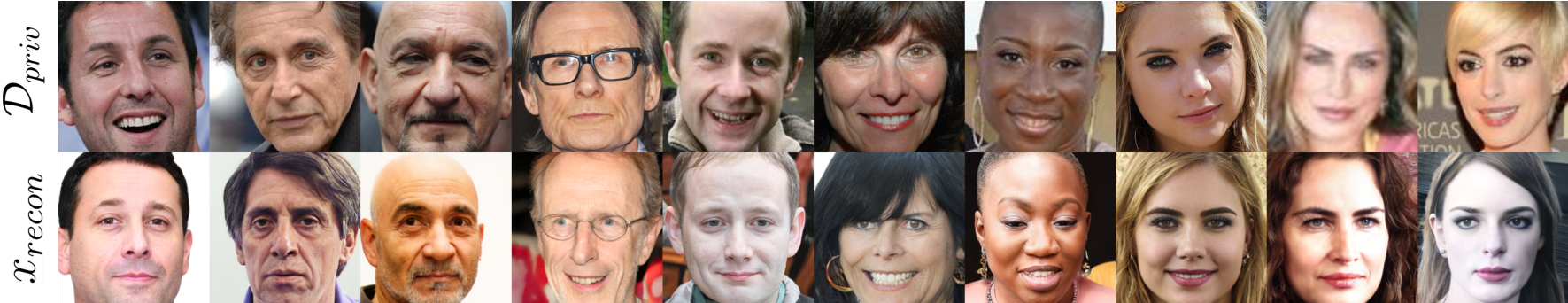}
    \caption{Qualitative results on the Facescrub dataset using LLaVA-v1.6-7B model with our SMI-AW and $\mathcal{L}_{LOM}$. The first row shows images from the private training dataset, while the second row presents the reconstructed images corresponding to each individual in the first row. The visual similarity between the original and reconstructed images demonstrates the effectiveness of our inversion method in recovering private training data.
    \textbf{More reconstructed images can be found in Supp.}
    }
    \label{fig:facescrub_qualitative}
\end{figure*}


Figure~\ref{fig:facescrub_qualitative} shows qualitative results demonstrating the effectiveness of our method. Using SMI-AW with $\mathcal{L}_{LOM}$, the reconstructed images from the LLaVA-v1.6-7B model (second row) closely resemble the corresponding identities in $\mathcal{D}_{priv}$ (first row). This strong visual similarity highlights the ability of our model inversion approach to recover identifiable features from the training data.
\textbf{More reconstructed images of other models/datasets can be found in Supp.}

\begin{table}[]
    \centering
    \caption{Human evaluation results. We evaluate our SMI-AW  method using $\mathcal{L}_{LOM}$, the private datasets $\mathcal{D}_{priv}$ are FaceScrub, CelebA, and StanfordDogs.}
    \begin{adjustbox}{width=0.8\linewidth}
        \begin{tabular}{llc}
        \hline
        VLM & $\mathcal{D}_{priv}$ &$AttAcc_{H} \uparrow$  \\ \hline
        LLaVA-v1.6-7B & \multirow{4}{*}{Facescrub} & 56.93\% \\
        MiniGPT-v2 &  & 57.22\%   \\
        Qwen2.5-VL-7B &  & 54.48\%  \\ 
        InternVL2.5-8B &  & 53.42\%  \\ \hline
        \multirow{2}{*}{LLaVA-v1.6-7B} & CelebA &  61.21\%  \\ \cline{2-3}
         & StanfordDogs &  55.42\%  \\ \hline        
        \end{tabular}
        \end{adjustbox}
        \label{tab:human}
\end{table}

\subsection{Human Evaluation}

We further conduct human evaluation on reconstructed images using three datasets Facescrub, CelebA, StanfordDogs. Each user study involves 4,240 participants for the FaceScrub dataset, 8,000 participants for the CelebA dataset, and 960 participants for the StanfordDogs dataset.
The results show that 53.42\% to 61.21\% of the reconstructed samples are deemed successful attacks, i.e., human annotators recognize the generated images as depicting the same identity as those in the private image set (see Table \ref{tab:human}). This highlights the alarming potential of such inversion attacks to compromise sensitive identity information. {\bf See details of human evaluation in Supp.}

\subsection{Evaluation with Publicly Released VLM}

In the experiments above, we fine-tuned the target model using a private training dataset following prior MI work on conventional DNNs\citep{chen2021knowledge,nguyen2023re, struppek2022plug,qiu2024closer}. In this section, we extend our analysis to the publicly available LLaVA-v1.6-7B model, aiming to reconstruct potential training images directly from it.

\begin{figure}[t!]
    \centering
    
\begin{subfigure}[b]{0.4\linewidth}
        \centering        
        \includegraphics[width=\linewidth]{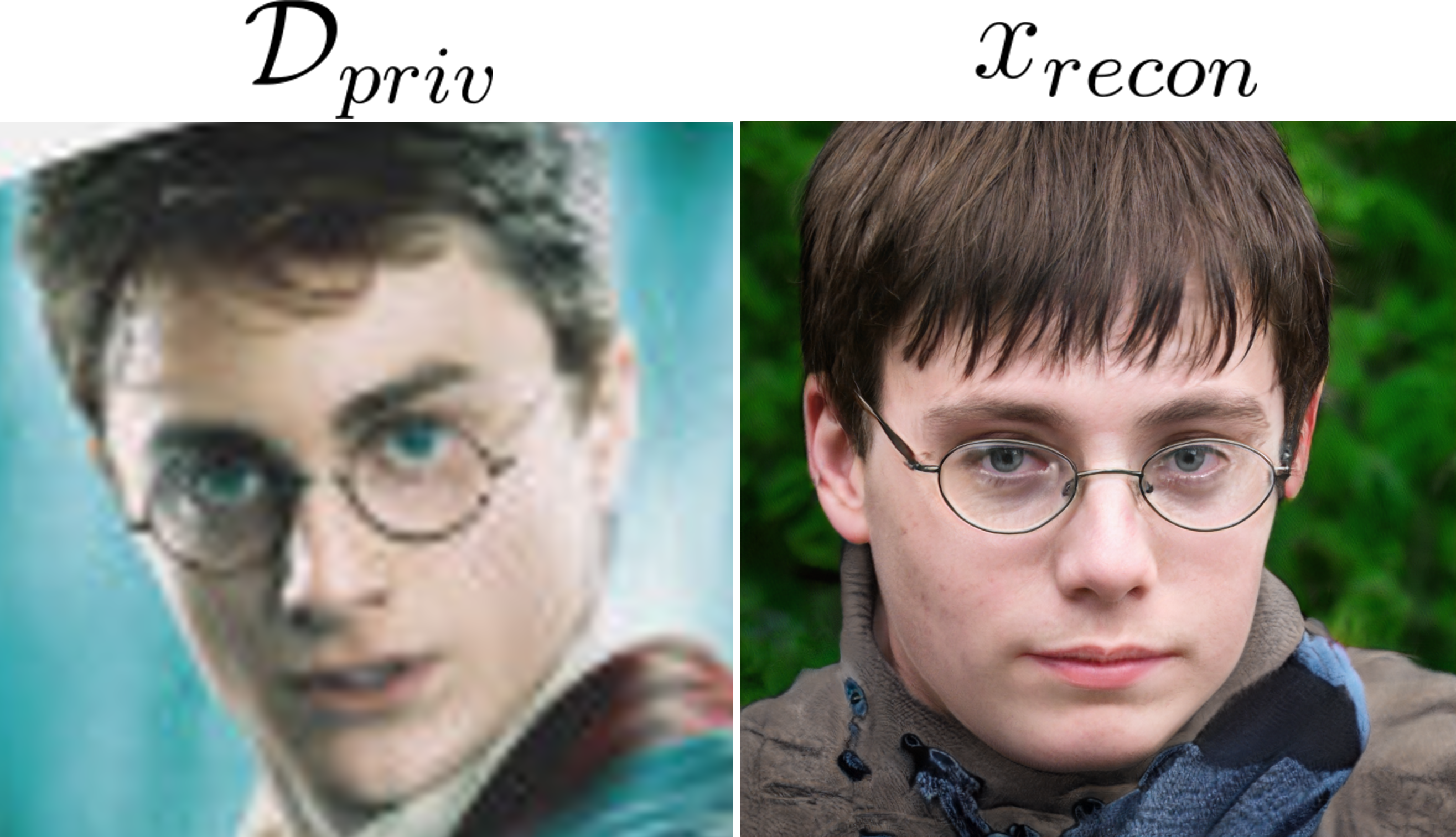}  
        \caption{Harry Potter}
        \label{fig:harry_potter}
    \end{subfigure}
    \hspace{0.5cm}
    \begin{subfigure}[b]{0.4\linewidth}
        \centering
        \includegraphics[width=\linewidth]{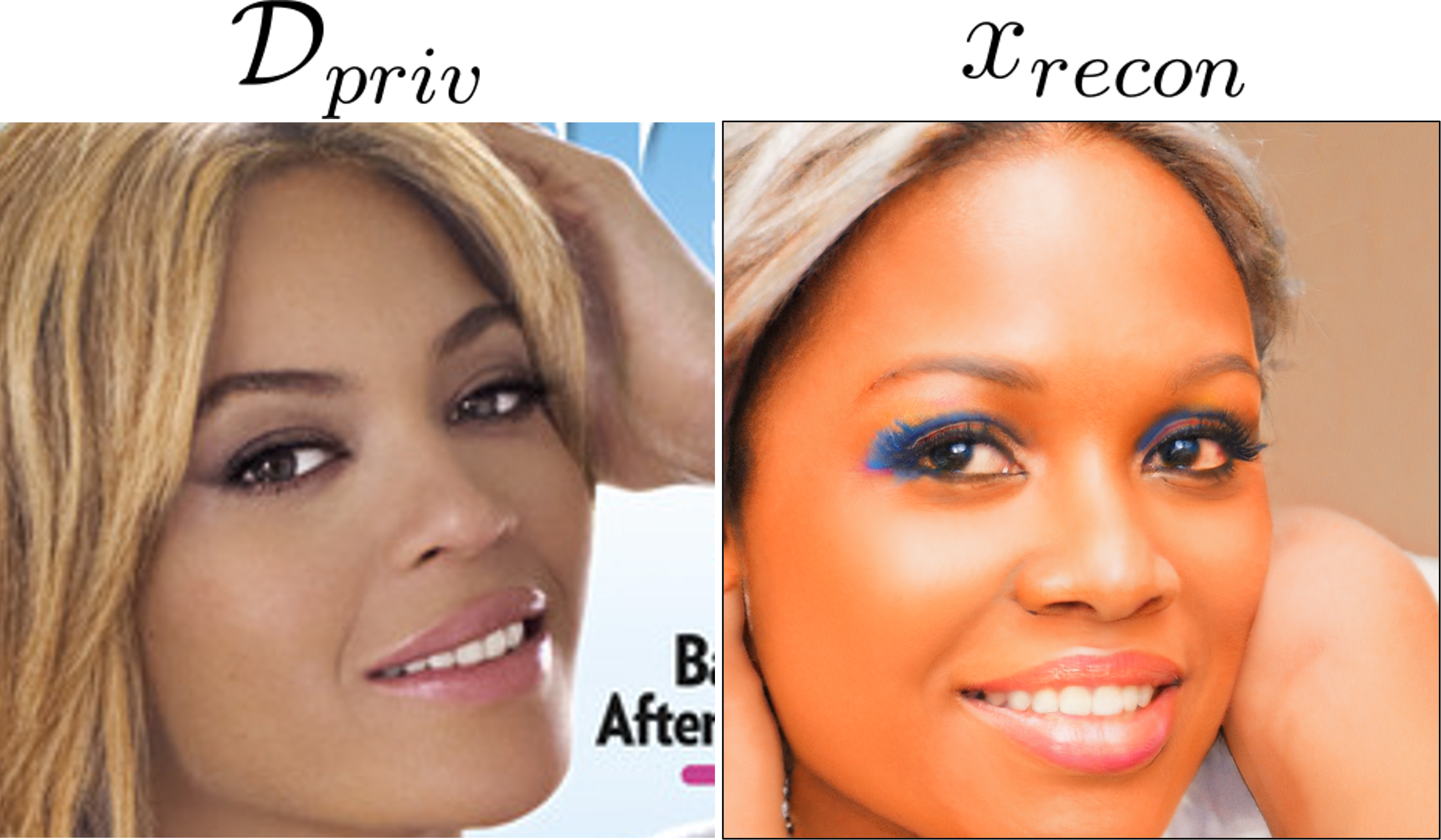}  
        \caption{Beyoncé}
        \label{fig:Beyoncé}
    \end{subfigure} 
    
\begin{subfigure}[b]{0.4\linewidth}
        \centering        
        \includegraphics[width=\linewidth]{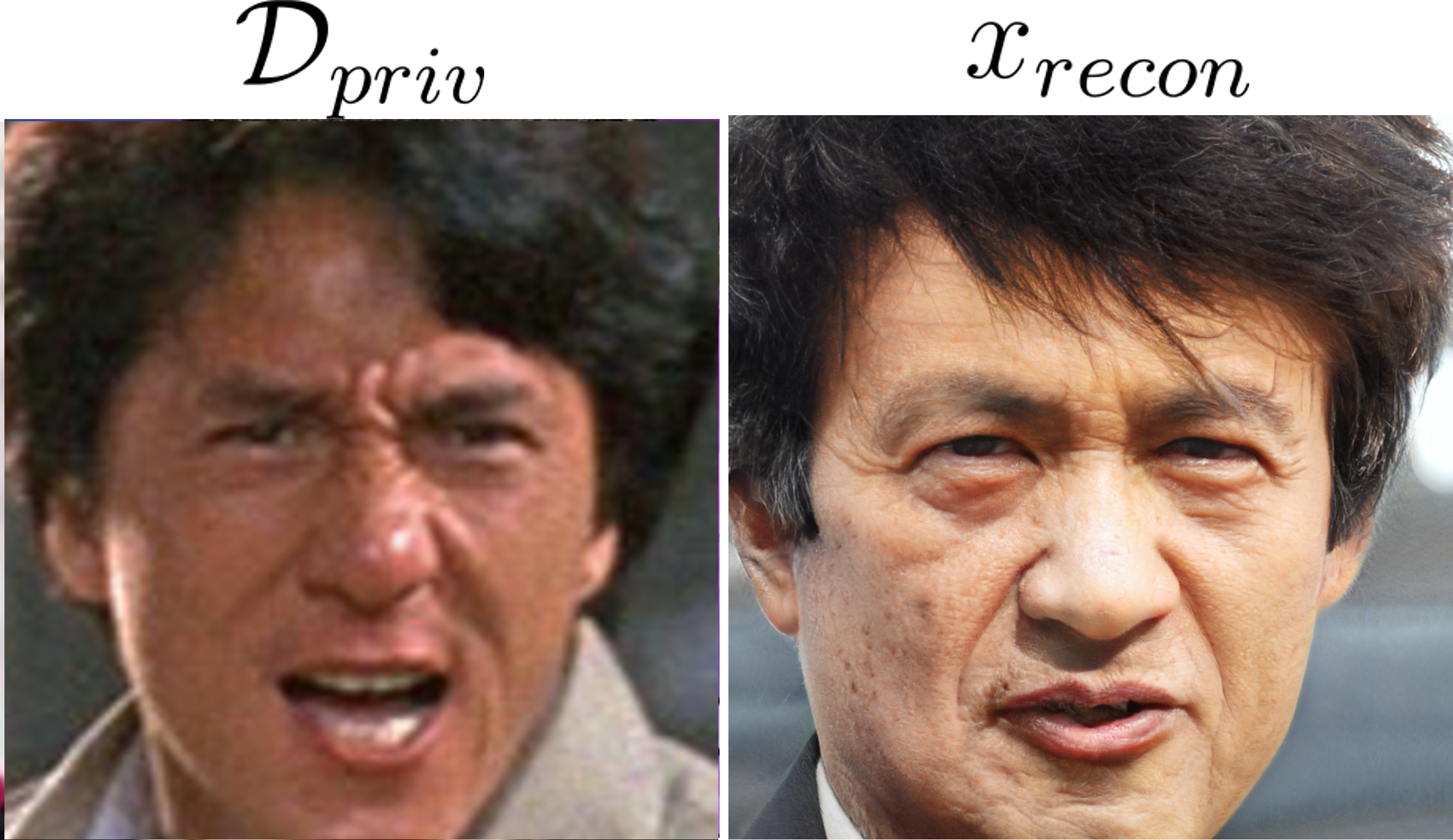}  
        \caption{Jackie Chan}
        \label{fig:Jackie}
    \end{subfigure}
    \hspace{0.5cm}
    \begin{subfigure}[b]{0.4\linewidth}
        \centering
        \includegraphics[width=\linewidth]{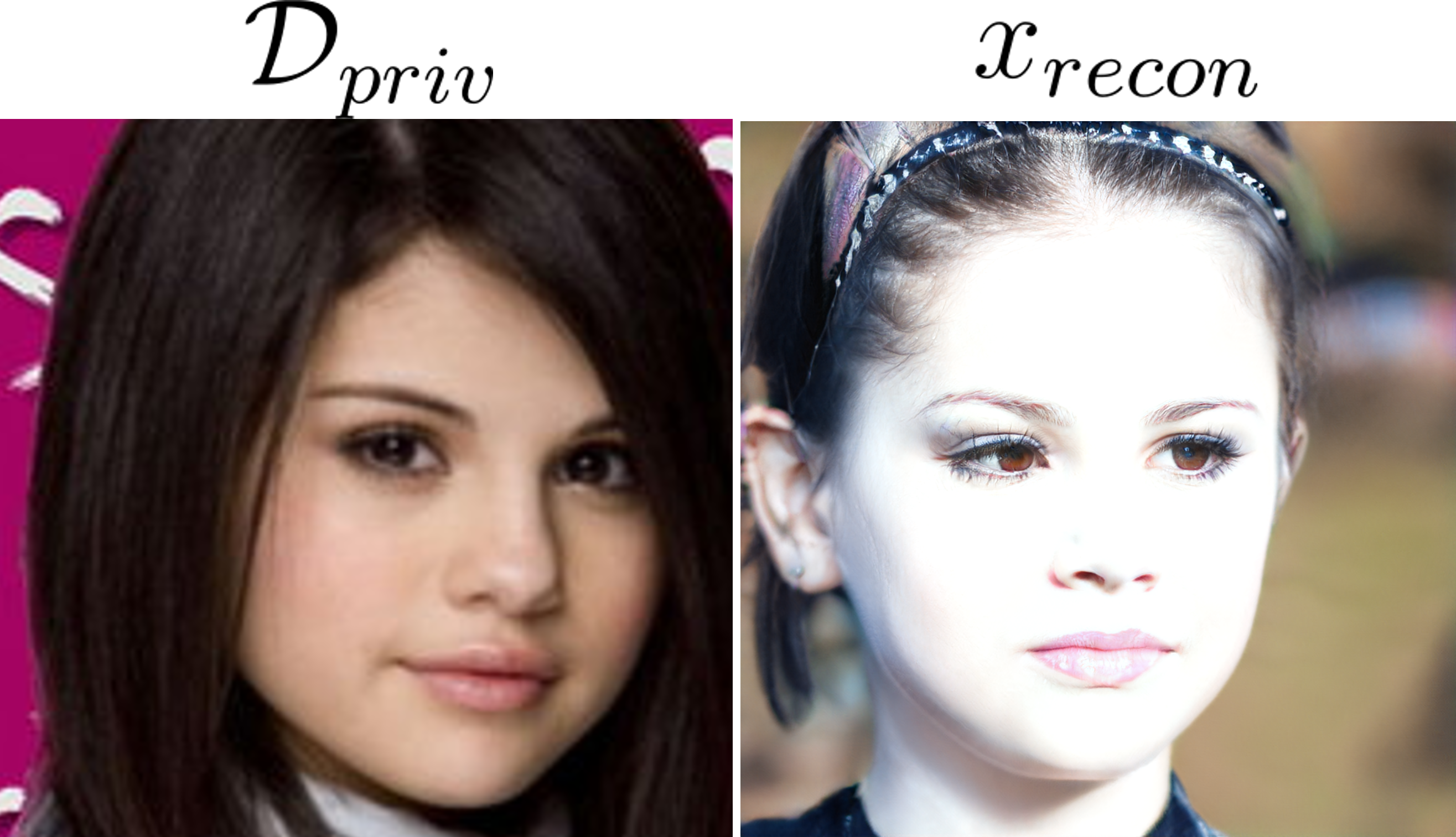}  
        \caption{Selena Gomez}
        \label{fig:Obama}
    \end{subfigure}    

    \caption{We reconstruct images of celebrities 
     from the pre-trained LLaVA-v1.6-7B model. We use SMI-AW with $\mathcal{L}_{LOM}$ to reconstruct images. For each pair, the left image shows a training image in $\mathcal{D}_{priv}$, while the right image presents the reconstruction $x_{recon}$ obtained via our model inversion attack. This result illustrates that the pre-trained VLM is vulnerable to training data leakage through model inversion. \textbf{More results can be found in Supp.}}
    \label{fig:pretrained_model}
\end{figure}

Figure~\ref{fig:pretrained_model} shows the results of our best setup of MI attack, SMI-AW using the logit maximization loss. The target is to reconstruct images of some identities that appear in the training dataset of the LLaVA-v1.6-7B model.
We present four image pairs: in each pair, the left image is a training sample of an identity, while the right image shows the corresponding reconstruction generated by the publicly available model.
The visual similarity between the pairs indicates that the pre-trained VLM may reveal  identifiable information from its training data, exposing its MI vulnerability.
\textbf{More results can be found in Supp.}

\section{Conclusion}
This study pioneers the investigation of MI attacks on VLMs, demonstrating for the first time their susceptibility to leaking private visual training data.
Our novel token-based and sequence-based inversion strategies reveal significant privacy risks across state-of-the-art and publicly available VLMs. Particularly, our proposed Sequence-based Model Inversion with Adaptive Token Weighting (SMI-AW) achieves an attack accuracy of 61.21\%. 
These findings underscore the privacy concerns as VLMs become more prevalent in real-world applications.
{\bf Additional analysis, limitation and broader impact are included in Supp.}

\noindent \textbf{Acknowledgements.}
This research is supported by the National Research Foundation, Singapore under its AI Singapore Programmes (AISG Award No.: AISG2-TC-2022-007); The Agency for Science, Technology and Research (A*STAR) under its MTC Programmatic Funds (Grant No. M23L7b0021). This research is supported by the National Research Foundation, Singapore and Infocomm Media Development Authority under its Trust Tech Funding Initiative. Any opinions, findings and conclusions or recommendations expressed in this material are those of the author(s) and do not reflect the views of National Research Foundation, Singapore and Infocomm Media Development Authority. The work is sponsored by the SUTD Decentralised Gap Funding Grant.

{
    \small
    \bibliographystyle{ieeenat_fullname}
    \bibliography{main}
}
\clearpage

\unhidefromtoc


\setcounter{figure}{0} 
\setcounter{table}{0} 
\setcounter{section}{0} 
\renewcommand{\thefigure}{S.\arabic{figure}}
\renewcommand{\thetable}{S.\arabic{table}}

{\centering
\Large\textbf{Supplementary material}
}

In this supplementary material, we provide additional experiments, analysis, ablation study, and details that are required to reproduce our results. These are not included in the main paper due to space limitations.

\tableofcontents


\section{Research Reproducibility Details
}

\subsection{Hyperparameters}
To fine-tune the VLMs, we follow the standard hyperparameters provided in the official implementations of LLaVA-v1.6-Vicuna-7B\footnote{\url{https://github.com/haotian-liu/LLaVA}} \citep{liu2024llavanext}, Qwen2.5-VL-7B\footnote{\url{https://github.com/QwenLM/Qwen2.5-VL}} \citep{Qwen2.5-VL}, MiniGPT-v2\footnote{\url{https://github.com/Vision-CAIR/MiniGPT-4}} \citep{chen2023minigptv2}, and InternVL2.5 \footnote{\url{https://github.com/OpenGVLab/InternVL}} \citep{chen2024internvl,chen2024expanding}. Fine-tuning is conducted on the VQA-FaceScrub, VQA-CelebA, and VQA-StanfordDogs datasets.

For the attacks, we use $N = 70$ inversion steps for all experiments.
The inversion update rate $\beta=0.05$.

To compute the regularization term $f_{reg}$ in Eqn.~\ref{eq:LOM}, we follow \citep{nguyen2023re} by using 2,000 images from a public dataset $\mathcal{D}_{pub}$ to estimate the mean and variance of the penultimate layer activations of the VLMs.

\subsection{Computational Resources}

All experiments were conducted on NVIDIA RTX A6000 Ada GPUs running Ubuntu 20.04.2 LTS, equipped with AMD Ryzen Threadripper PRO 5975WX 32-core processors.
The environment setup for each model is provided in the official implementations of the VLMs, including:
LLaVA-v1.6-Vicuna-7B \citep{liu2024llavanext},
Qwen2.5-VL-7B \citep{Qwen2.5-VL}, 
MiniGPT-v2 \citep{chen2023minigptv2}, and InternVL2.5 \citep{chen2024internvl,chen2024expanding}.

To evaluate $AttAcc_M$, we strictly follow the protocol in \citep{ho2025revisitingmodelinversionevaluation}, using the Gemini 2.0 Flash API.
In total, we evaluate nearly 100,000 MI-reconstructed images for our main experiments (main paper).

\section{Additional results}
\subsection{Extended Evaluation on Publicly Released VLM}

In this section, we extend our analysis to the publicly available LLaVA-v1.6-7B model \citep{liu2024llavanext} and MiniGPTv2 \citep{chen2023minigptv2}, aiming to reconstruct training images from accessing the model only.

\begin{figure}[t!]
    \centering
    
    \begin{subfigure}[b]{0.6\columnwidth}
        \centering
        \includegraphics[width=0.96\columnwidth]
    {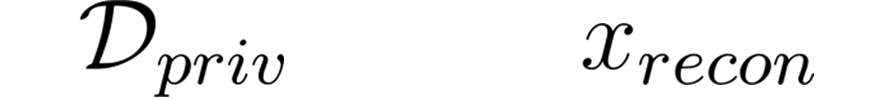}  
        \includegraphics[width=\columnwidth]{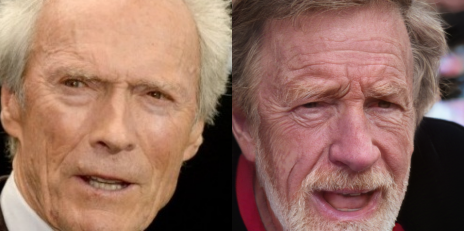}
        \caption{Clint Eastwood}
    \end{subfigure}    
    
    \begin{subfigure}[b]{0.6\columnwidth}
        \centering
        \includegraphics[width=\columnwidth]{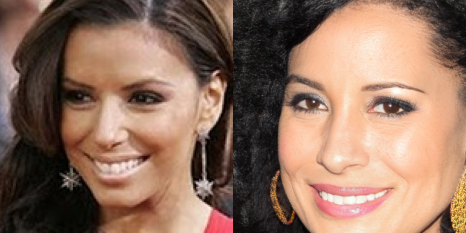}
        \caption{Eva Longoria}
    \end{subfigure}
    
    \begin{subfigure}[b]{0.6\columnwidth}
        \centering
        \includegraphics[width=\columnwidth]{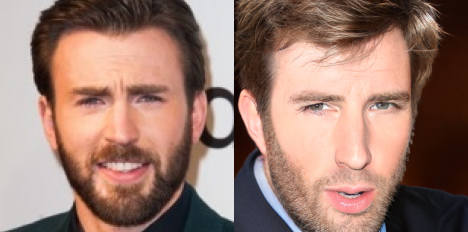}
        \caption{Chris Evans}
    \end{subfigure}
    
    \begin{subfigure}[b]{0.6\columnwidth}
        \centering
        \includegraphics[width=\columnwidth]{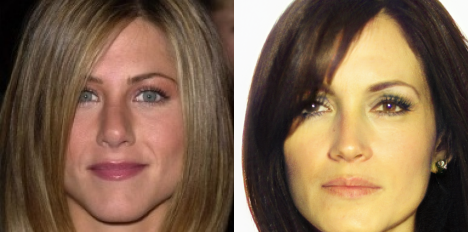}
        \caption{Jennifer Aniston}
    \end{subfigure}

    \begin{subfigure}[b]{0.6\columnwidth}
        \centering
        \includegraphics[width=\columnwidth]{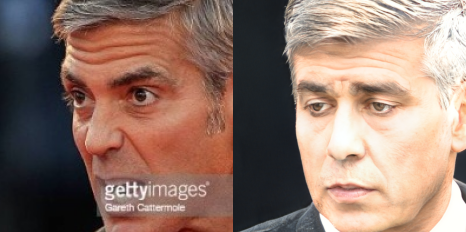}
        \caption{George Clooney}
    \end{subfigure}

    \caption{Reconstructed images 
    using our SMI-AW with $\mathcal{L}_{LOM}$ on the publicly available LLaVA-v1.6-7B model. Each pair consists of a reconstructed image (right) and a corresponding training image (left) in the training dataset of LLaVA-v1.6-7B model. 
    We emphasize that the training dataset is fully unknown and inaccessible for the inversion attack.
    The strong similarity suggests the pre-trained VLM may leak identifiable training data, exposing it to model inversion attacks.
    }
    \label{fig:llava_pretrained_model}
\end{figure}

\begin{figure}[t!]
    \centering    
    
     \begin{subfigure}[b]{0.6\columnwidth}
        \centering        
        \includegraphics[width=0.96\columnwidth]
    {figures_supp/x_recons_priv_v2.png} 
        \includegraphics[width=\columnwidth]{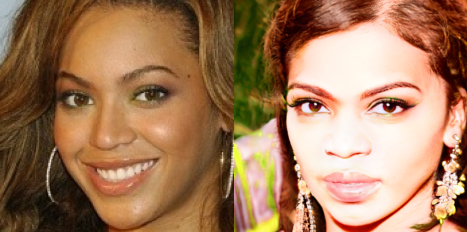}
        \caption{Beyoncé}
    \end{subfigure}    
    
     \begin{subfigure}[b]{0.6\columnwidth}
        \centering 
        \includegraphics[width=\columnwidth]{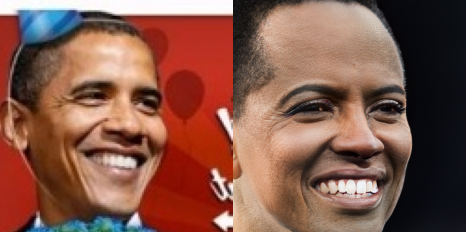}
        \caption{Barack Obama}
    \end{subfigure}    
    
     \begin{subfigure}[b]{0.6\columnwidth}
        \centering 
        \includegraphics[width=\columnwidth]{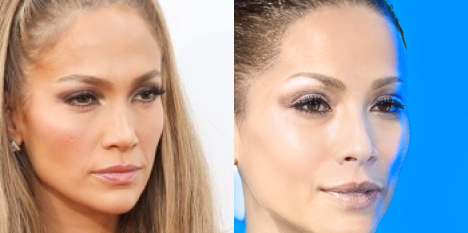}
        \caption{Jennifer Lopez}
    \end{subfigure}  
    
     \begin{subfigure}[b]{0.6\columnwidth}
        \centering
        \includegraphics[width=\columnwidth]{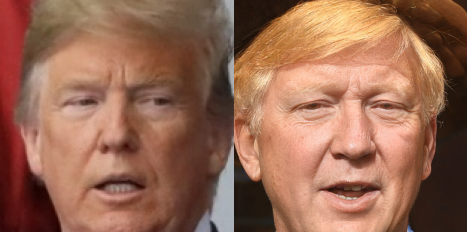}
        \caption{Donald Trump}
    \end{subfigure}

     \begin{subfigure}[b]{0.6\columnwidth}
        \centering
        \includegraphics[width=\columnwidth]{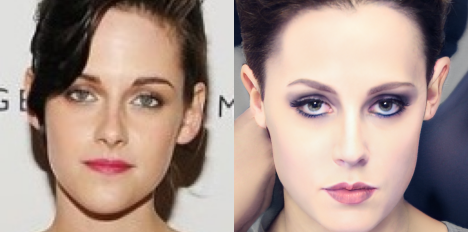}
        \caption{Kristen Stewart}
    \end{subfigure}

    \caption{Reconstructed images   
    using our SMI-AW with $\mathcal{L}_{LOM}$ on the publicly available MiniGPTv2 model. Each pair consists of a reconstructed image (right) and a corresponding training image (left) in the training dataset of MiniGPTv2 model. 
    We emphasize that the training dataset is fully unknown and inaccessible for the inversion attack.
    The strong similarity suggests the pre-trained VLM may leak identifiable training data, exposing it to model inversion attacks.
    }
    \label{fig:minigpt_pretrained_model}
\end{figure}

Figure \ref{fig:llava_pretrained_model}
and Figure \ref{fig:minigpt_pretrained_model}
show the results of our best setup of MI attack, SMI-AW using the logit maximization loss $\mathcal{L}_{LOM}$. The target is to reconstruct images of celebrities that appear in the training dataset of the LLaVA-v1.6-7B and MiniGPTv2 model. To reconstructed images from the model, we use the  textual input $t$ = ``What is the person's name in the image? Return only their name'' and the target textual answer is a celebrity's name, i.e $y$ = ``Beyoncé''.

We visualize  image pairs: in each pair, the right image is the reconstruction generated from the publicly available model, and the left image is a training image of an individual. 
We emphasize that the training dataset is fully unknown and inaccessible for the inversion attack.
The  visual similarity between the pairs indicates that the pre-trained VLM may reveal identifiable information from its training data, exposing a vulnerability to model inversion attacks.

\subsection{Additional Qualitative Results}

\begin{figure} [t!]
    \centering    
    \includegraphics[width=0.43\linewidth]{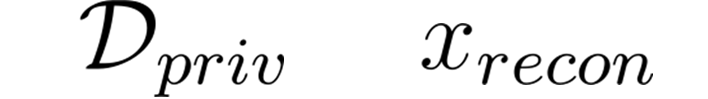}    \hspace{0.5cm}
    \includegraphics[width=0.43\linewidth]
    {figures_supp/x_recons_priv.png}  
    
    \includegraphics[width=0.43\linewidth]{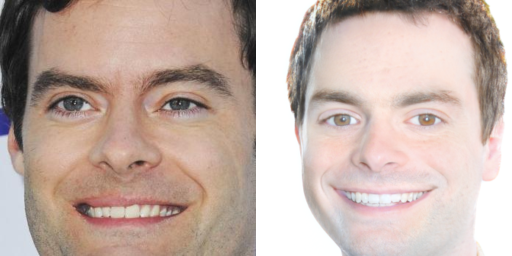}    \hspace{0.5cm}
    \includegraphics[width=0.43\linewidth]
    {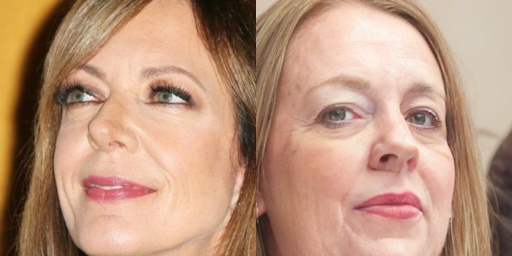}

    \includegraphics[width=0.43\linewidth]{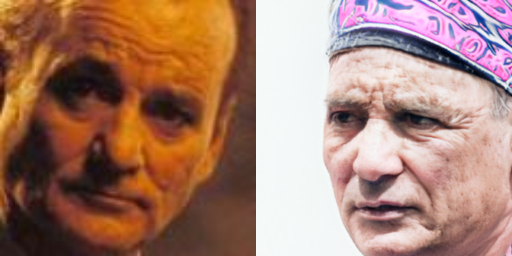}    \hspace{0.5cm}
    \includegraphics[width=0.43\linewidth]
    {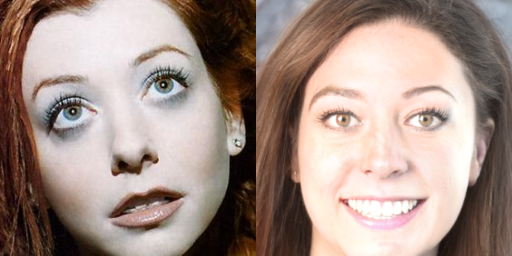}

    \includegraphics[width=0.43\linewidth]{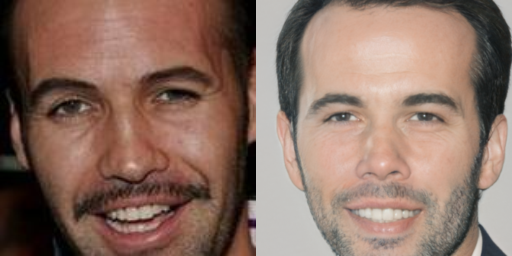}    \hspace{0.5cm}
    \includegraphics[width=0.43\linewidth]
    {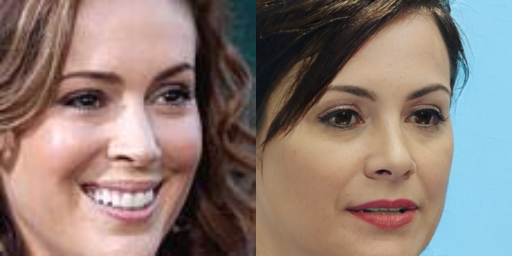}

    \includegraphics[width=0.43\linewidth]{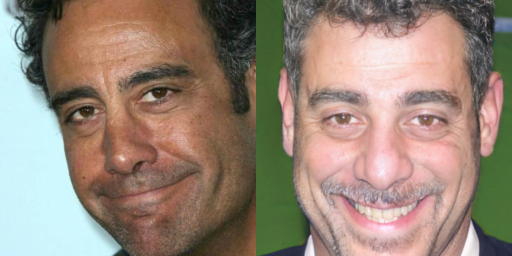}    \hspace{0.5cm}
    \includegraphics[width=0.43\linewidth]
    {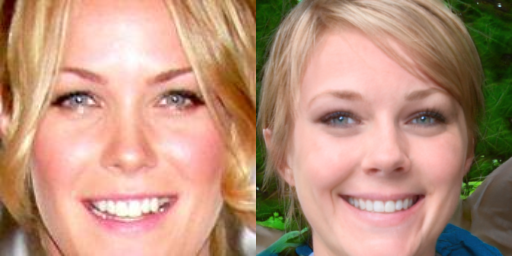}

    \includegraphics[width=0.43\linewidth]{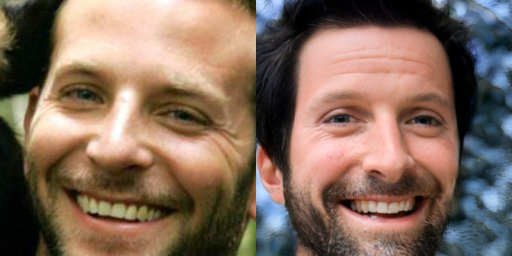}    \hspace{0.5cm}
    \includegraphics[width=0.43\linewidth]
    {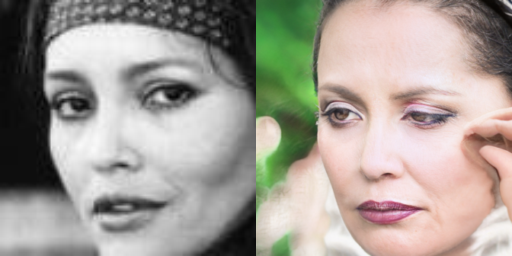}

    \includegraphics[width=0.43\linewidth]{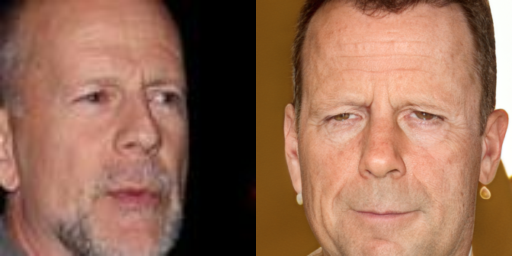}    \hspace{0.5cm}
    \includegraphics[width=0.43\linewidth]
    {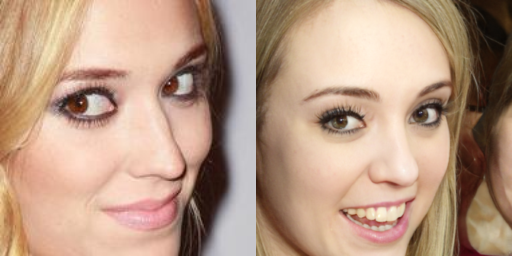}

    \includegraphics[width=0.43\linewidth]{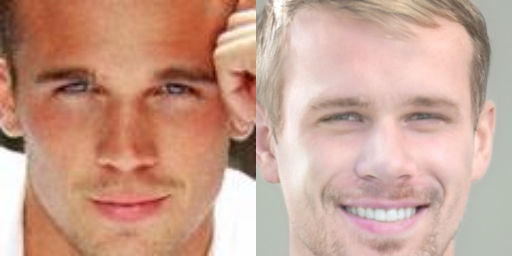}    \hspace{0.5cm}
    \includegraphics[width=0.43\linewidth]
    {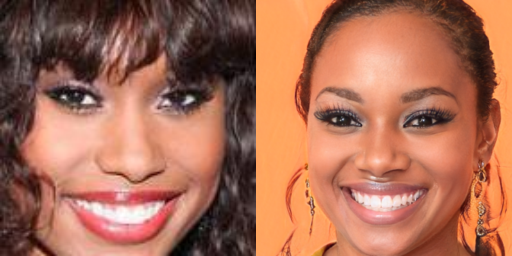}

    \includegraphics[width=0.43\linewidth]{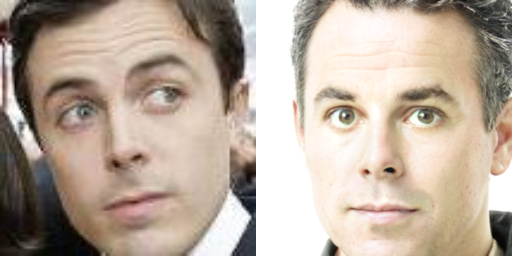}    \hspace{0.5cm}
    \includegraphics[width=0.43\linewidth]
    {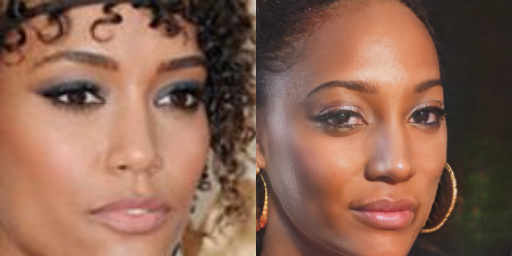}

    \includegraphics[width=0.43\linewidth]{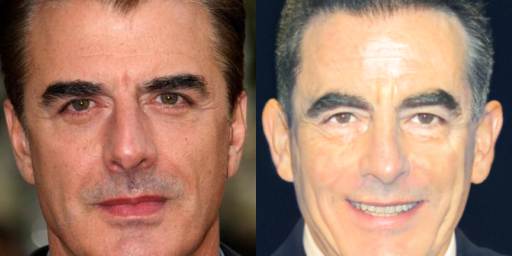}    \hspace{0.5cm}
    \includegraphics[width=0.43\linewidth]
    {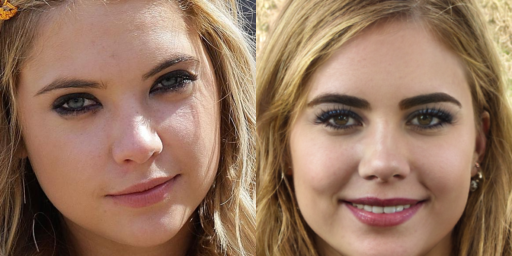}

    \includegraphics[width=0.43\linewidth]{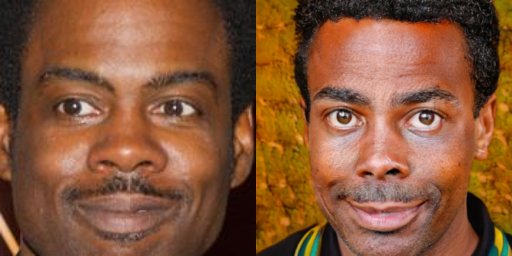}    \hspace{0.5cm}
    \includegraphics[width=0.43\linewidth]
    {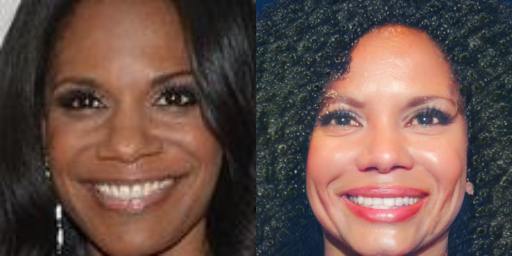}  

    \caption{Qualitative results on Facescrub dataset using the SMI-AW and $\mathcal{L}_{LOM}$, $M$ = LLaVA-v1.6-7B. For each pair, the left column shows images from the private training dataset, while the right column presents the reconstructed images corresponding to each individual in the left column. 
    }
    \label{fig:facescrub_llava}
\end{figure}

\begin{figure} [t!]
    \centering    
    \includegraphics[width=0.43\linewidth]{figures_supp/x_recons_priv.png}    \hspace{0.5cm}
    \includegraphics[width=0.43\linewidth]
    {figures_supp/x_recons_priv.png}

    \includegraphics[width=0.43\linewidth]{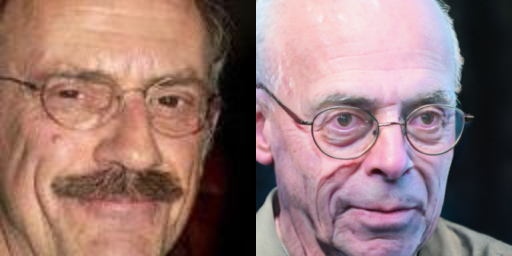}    \hspace{0.5cm}
    \includegraphics[width=0.43\linewidth]
    {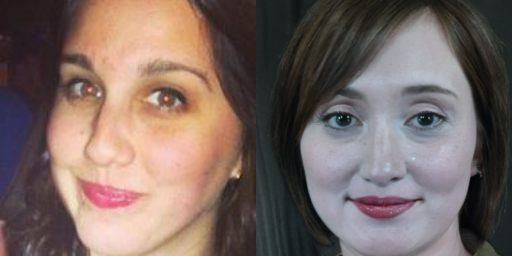}  
    
    \includegraphics[width=0.43\linewidth]{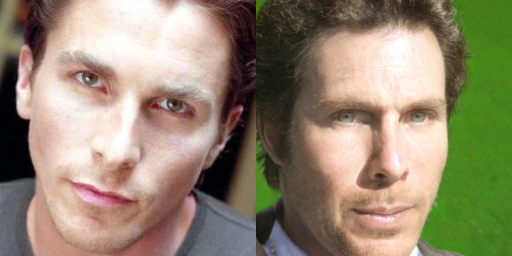}    \hspace{0.5cm}
    \includegraphics[width=0.43\linewidth]
    {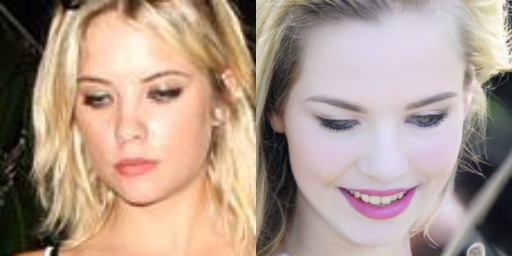}

    \includegraphics[width=0.43\linewidth]{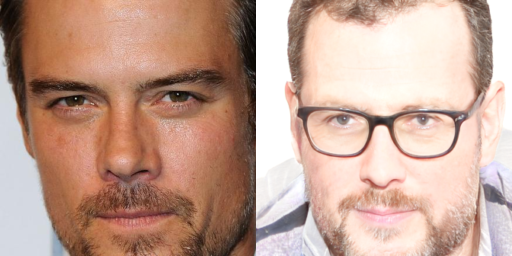}    \hspace{0.5cm}
    \includegraphics[width=0.43\linewidth]
    {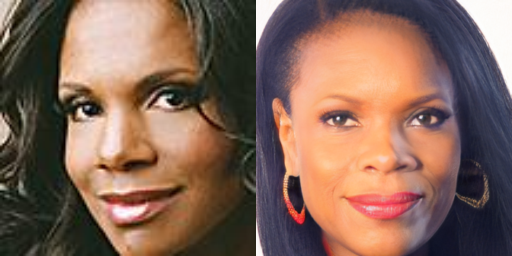}

    \includegraphics[width=0.43\linewidth]{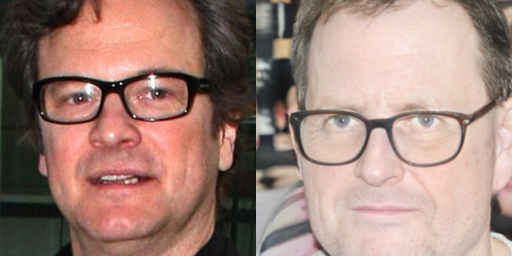}    \hspace{0.5cm}
    \includegraphics[width=0.43\linewidth]
    {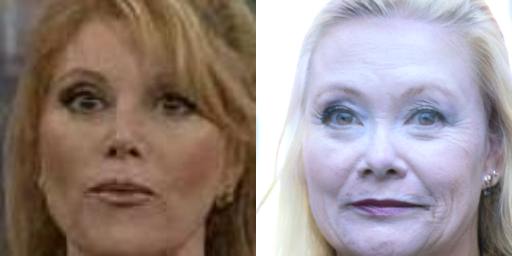}

    \includegraphics[width=0.43\linewidth]{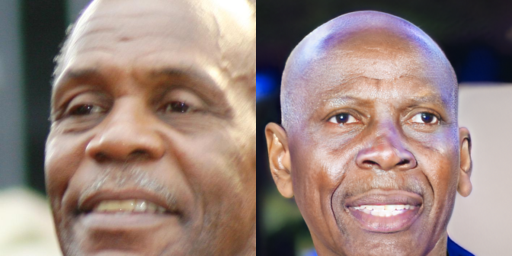}    \hspace{0.5cm}
    \includegraphics[width=0.43\linewidth]
    {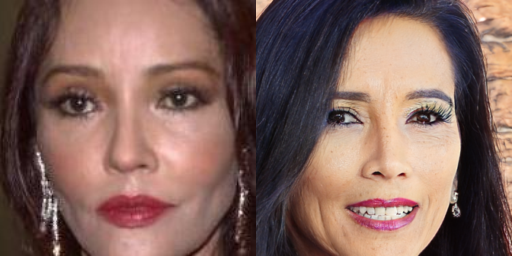}

    \includegraphics[width=0.43\linewidth]{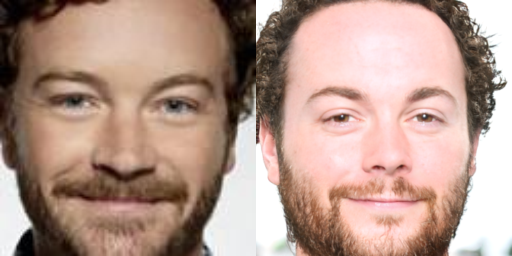}    \hspace{0.5cm}
    \includegraphics[width=0.43\linewidth]
    {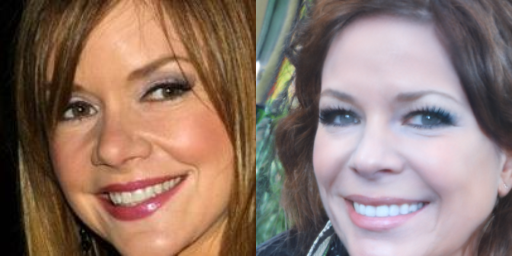}

    \includegraphics[width=0.43\linewidth]{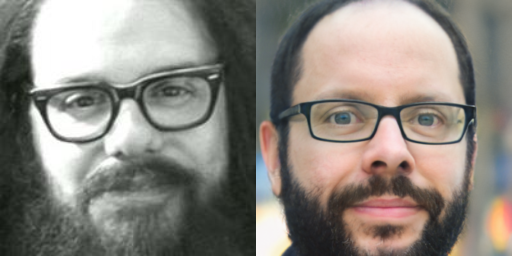}    \hspace{0.5cm}
    \includegraphics[width=0.43\linewidth]
    {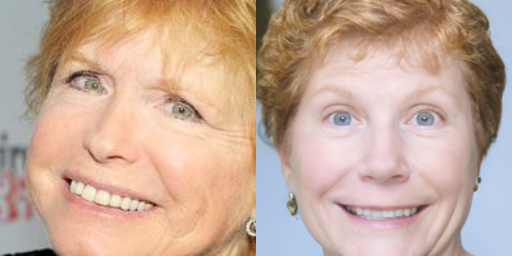}

    \includegraphics[width=0.43\linewidth]{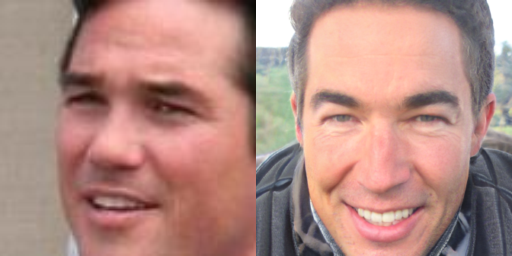}    \hspace{0.5cm}
    \includegraphics[width=0.43\linewidth]
    {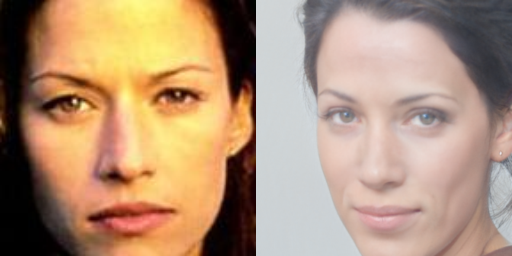}  

    \includegraphics[width=0.43\linewidth]{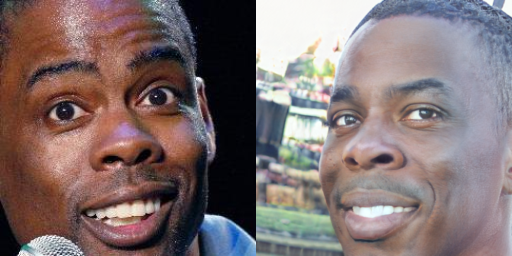}    \hspace{0.5cm}
    \includegraphics[width=0.43\linewidth]
    {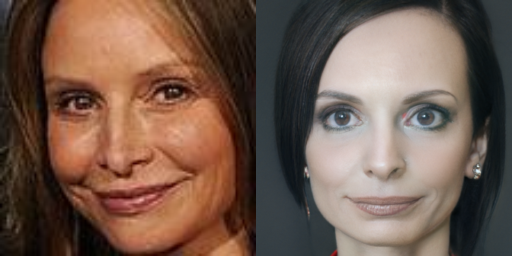}

    \includegraphics[width=0.43\linewidth]{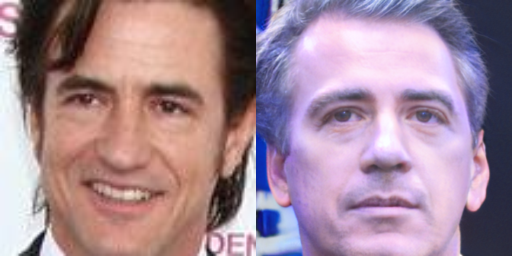}    \hspace{0.5cm}
    \includegraphics[width=0.43\linewidth]
    {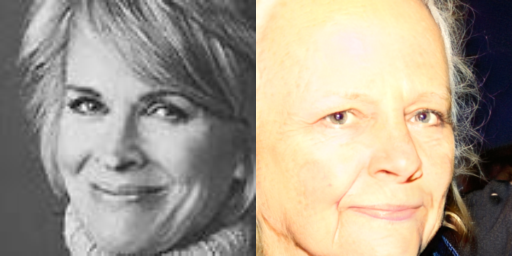}

    \caption{Qualitative results on Facescrub dataset using the SMI-AW and $\mathcal{L}_{LOM}$, $M$ = MiniGPT-v2. For each pair, the left column shows images from the private training dataset, while the right column presents the reconstructed images corresponding to each individual in the left column. 
    }
    \label{fig:facescrub_mingpt}
\end{figure}

\begin{figure} [t!]
    \centering    
    \includegraphics[width=0.43\linewidth]{figures_supp/x_recons_priv.png}    \hspace{0.5cm}
    \includegraphics[width=0.43\linewidth]
    {figures_supp/x_recons_priv.png}  
    
    \includegraphics[width=0.43\linewidth]{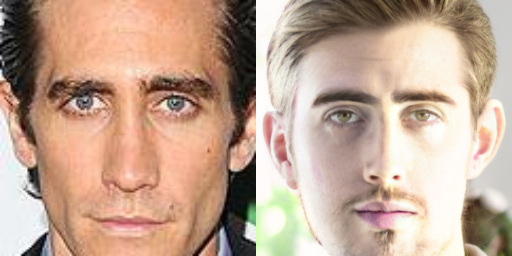}    \hspace{0.5cm}
    \includegraphics[width=0.43\linewidth]{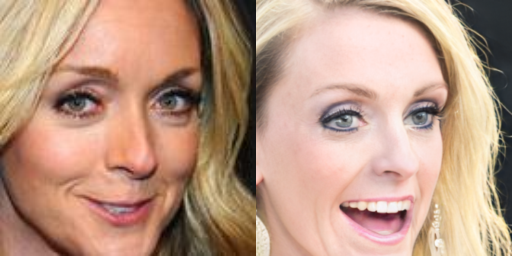}  
    
    \includegraphics[width=0.43\linewidth]{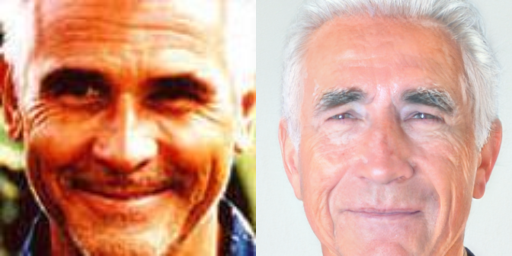}    \hspace{0.5cm}
    \includegraphics[width=0.43\linewidth]{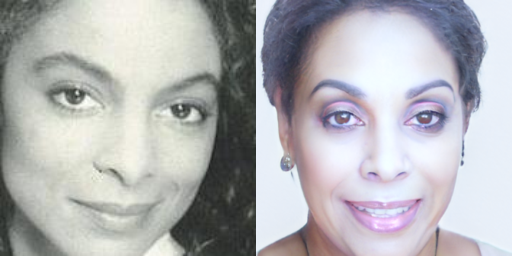}

    \includegraphics[width=0.43\linewidth]{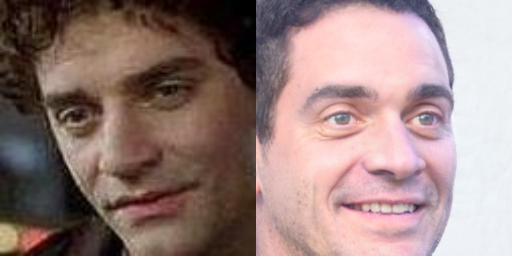}    \hspace{0.5cm}
    \includegraphics[width=0.43\linewidth]{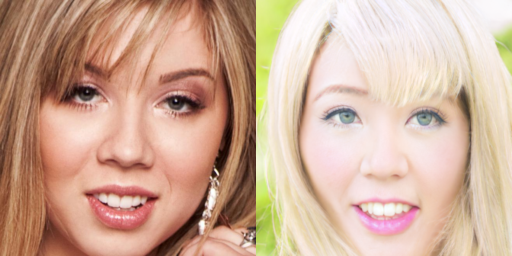}

    \includegraphics[width=0.43\linewidth]{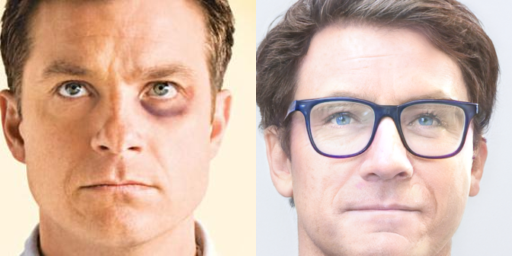}    \hspace{0.5cm}
    \includegraphics[width=0.43\linewidth]{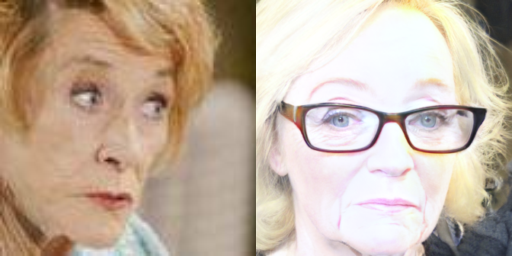}

    \includegraphics[width=0.43\linewidth]{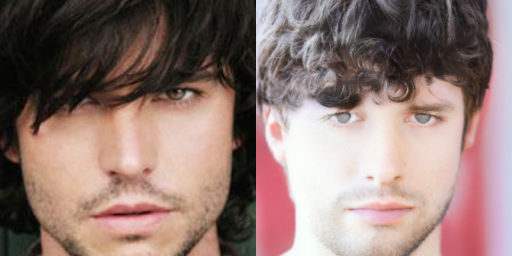}    \hspace{0.5cm}
    \includegraphics[width=0.43\linewidth]{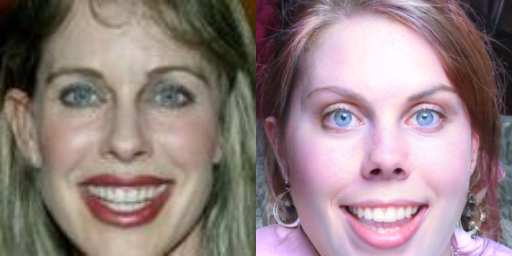}

    \includegraphics[width=0.43\linewidth]{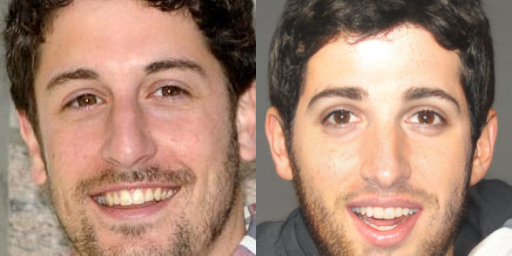}    \hspace{0.5cm}
    \includegraphics[width=0.43\linewidth]{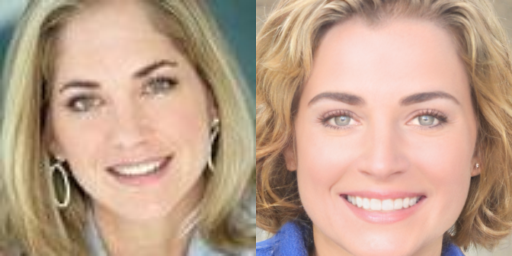}

    \includegraphics[width=0.43\linewidth]{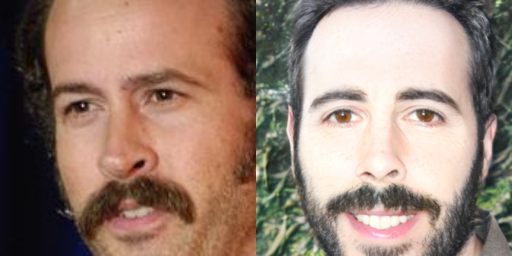}    \hspace{0.5cm}
    \includegraphics[width=0.43\linewidth]{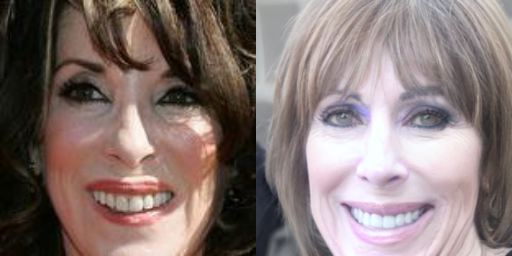}

    \includegraphics[width=0.43\linewidth]{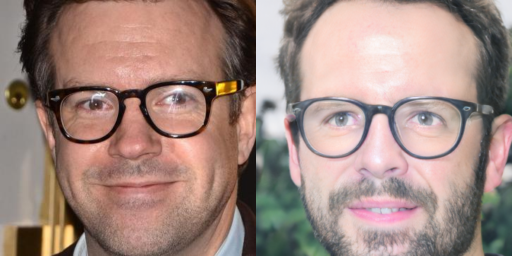}    \hspace{0.5cm}
    \includegraphics[width=0.43\linewidth]{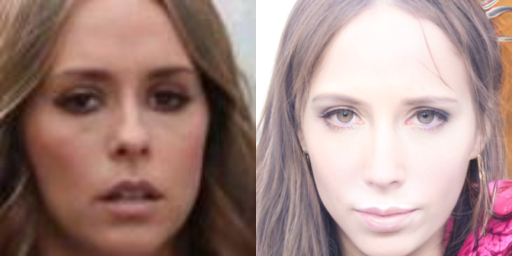}

    \includegraphics[width=0.43\linewidth]{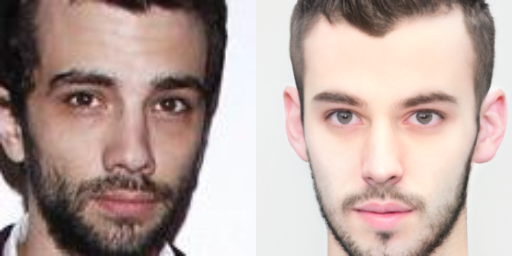}    \hspace{0.5cm}
    \includegraphics[width=0.43\linewidth]{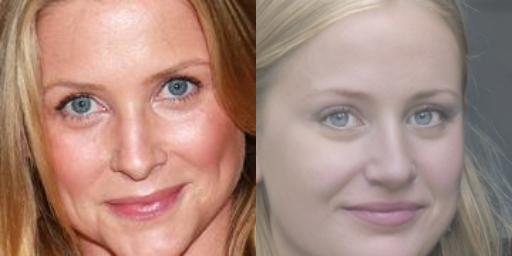}

    \includegraphics[width=0.43\linewidth]{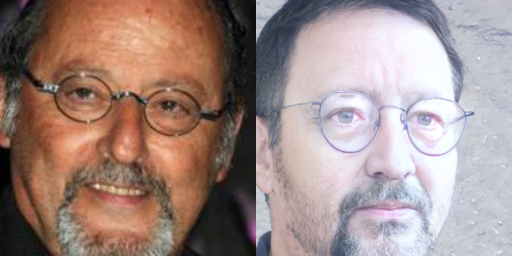}    \hspace{0.5cm}
    \includegraphics[width=0.43\linewidth]{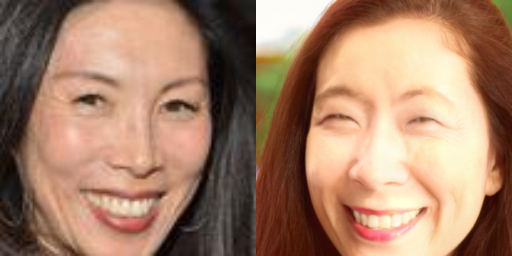}

    \caption{Qualitative results on Facescrub dataset using the SMI-AW and $\mathcal{L}_{LOM}$, $M$ = Qwen2.5-VL. For each pair, the left column shows images from the private training dataset, while the right column presents the reconstructed images corresponding to each individual in the left column. 
    }
    \label{fig:facescrub_qwen}
\end{figure}

\begin{figure} [t!]
    \centering    
    \includegraphics[width=0.43\linewidth]{figures_supp/x_recons_priv.png}    \hspace{0.5cm}
    \includegraphics[width=0.43\linewidth]
    {figures_supp/x_recons_priv.png}  
    
    \includegraphics[width=0.43\linewidth]{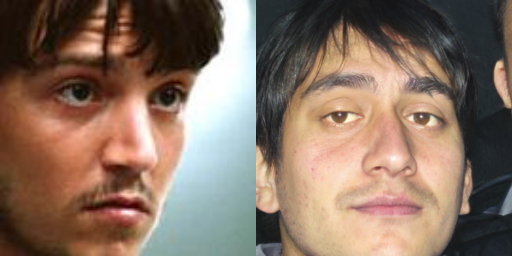}    \hspace{0.5cm}
    \includegraphics[width=0.43\linewidth]
    {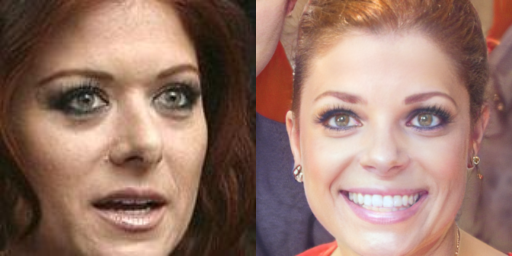}

    \includegraphics[width=0.43\linewidth]{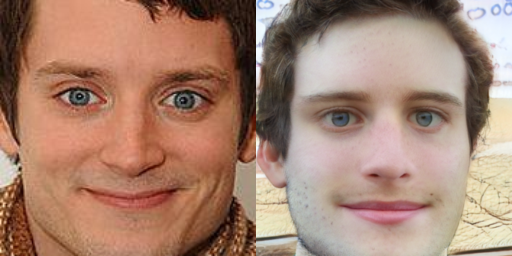}    \hspace{0.5cm}
    \includegraphics[width=0.43\linewidth]
    {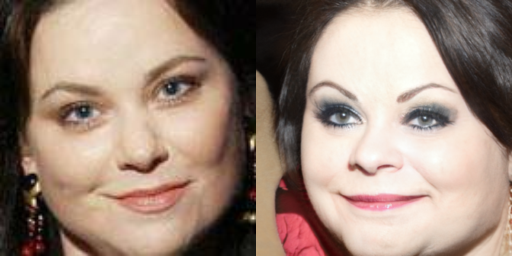} 
    
    \includegraphics[width=0.43\linewidth]{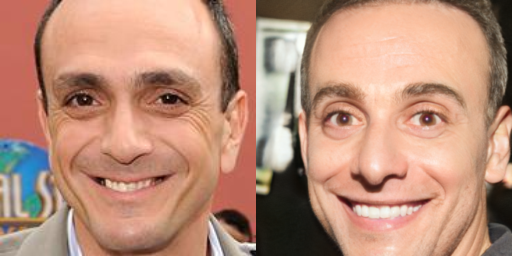}    \hspace{0.5cm}
    \includegraphics[width=0.43\linewidth]
    {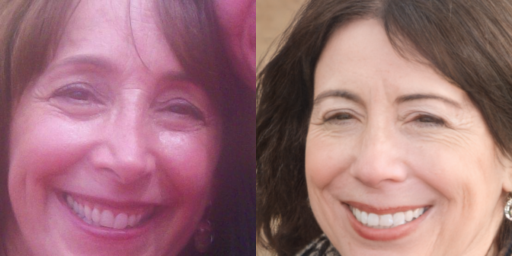} 
    
    \includegraphics[width=0.43\linewidth]{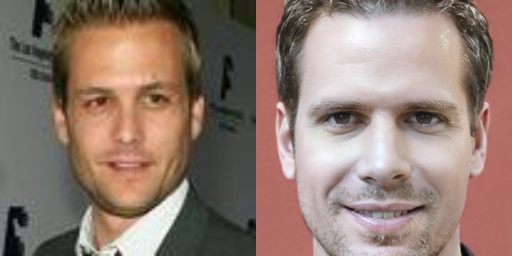}    \hspace{0.5cm}
    \includegraphics[width=0.43\linewidth]
    {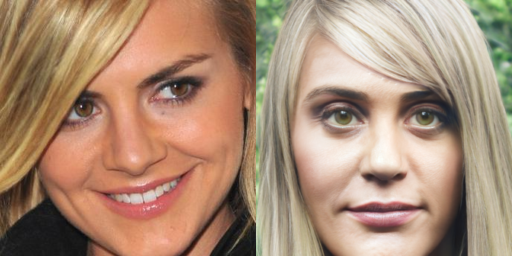} 
    
    \includegraphics[width=0.43\linewidth]{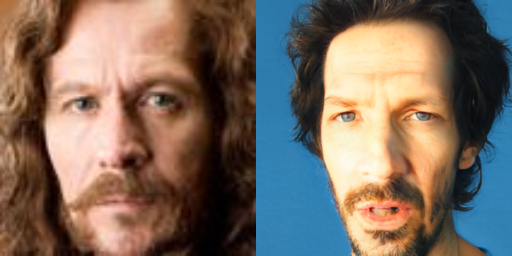}    \hspace{0.5cm}
    \includegraphics[width=0.43\linewidth]
    {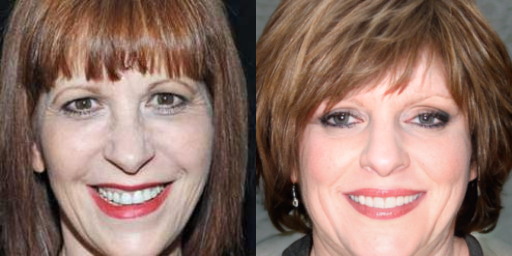} 
    
    \includegraphics[width=0.43\linewidth]{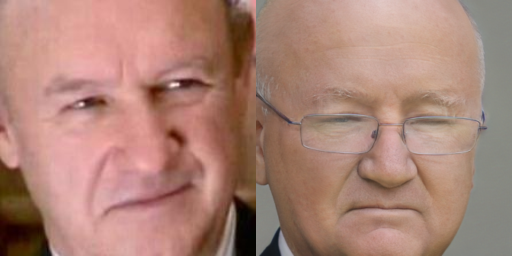}    \hspace{0.5cm}
    \includegraphics[width=0.43\linewidth]
    {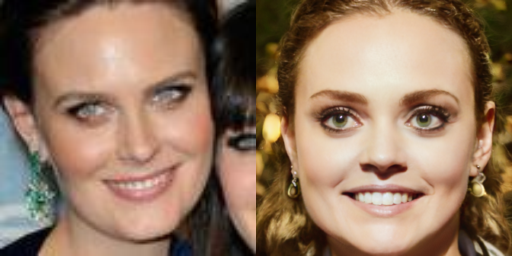} 
    
    \includegraphics[width=0.43\linewidth]{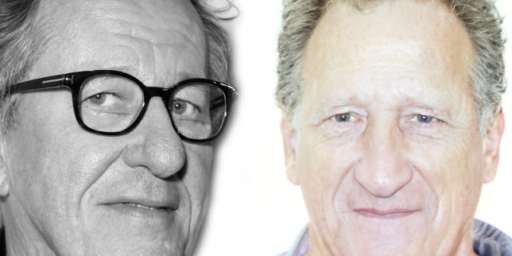}    \hspace{0.5cm}
    \includegraphics[width=0.43\linewidth]
    {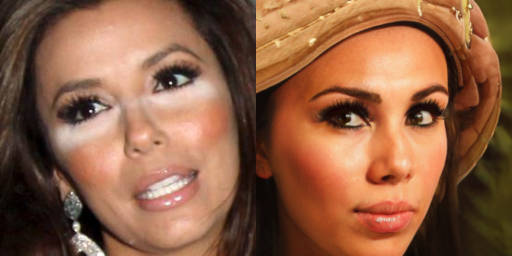} 
    
    \includegraphics[width=0.43\linewidth]{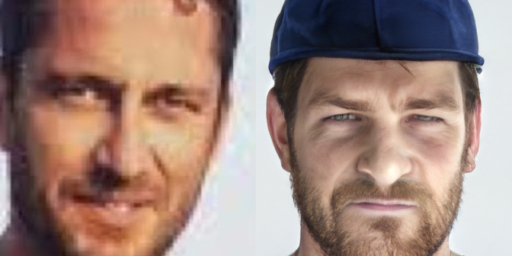}    \hspace{0.5cm}
    \includegraphics[width=0.43\linewidth]
    {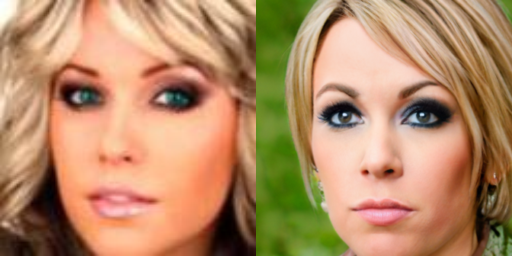} 
    
    \includegraphics[width=0.43\linewidth]{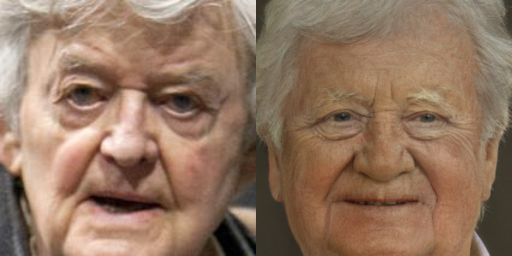}    \hspace{0.5cm}
    \includegraphics[width=0.43\linewidth]
    {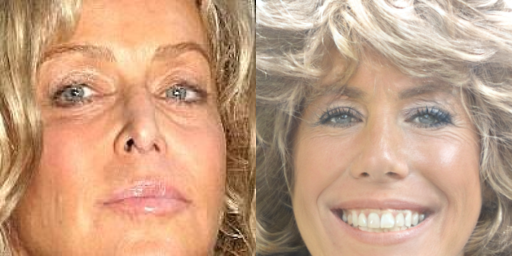} 
     
    \includegraphics[width=0.43\linewidth]{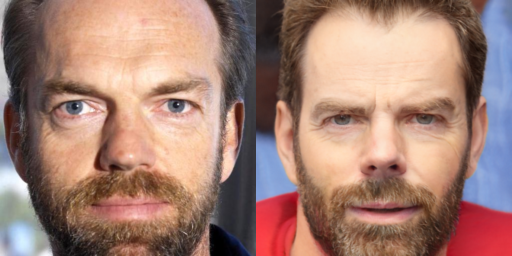}    \hspace{0.5cm}
    \includegraphics[width=0.43\linewidth]
    {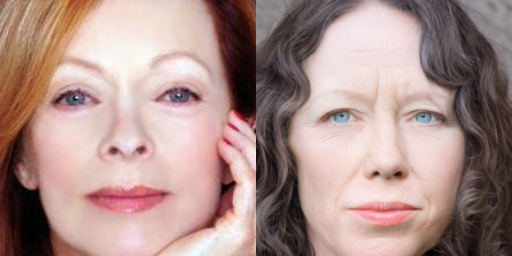}

    \caption{Qualitative results on Facescrub dataset using the SMI-AW and $\mathcal{L}_{LOM}$, $M$ = InternVL2.5. For each pair, the left column shows images from the private training dataset, while the right column presents the reconstructed images corresponding to each individual in the left column. 
    }
    \label{fig:facescrub_internvl}
\end{figure}

\begin{figure}[t!]
    \centering    
    \includegraphics[width=0.43\linewidth]{figures_supp/x_recons_priv.png}    \hspace{0.5cm}
    \includegraphics[width=0.43\linewidth]
    {figures_supp/x_recons_priv.png}  

    \includegraphics[width=0.43\linewidth]{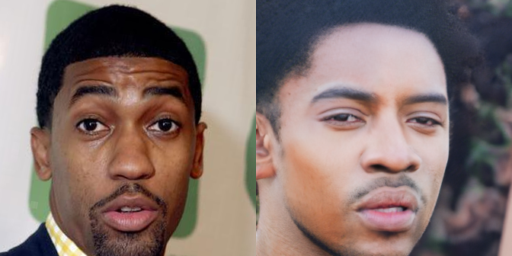}    \hspace{0.5cm}
     \includegraphics[width=0.43\linewidth]{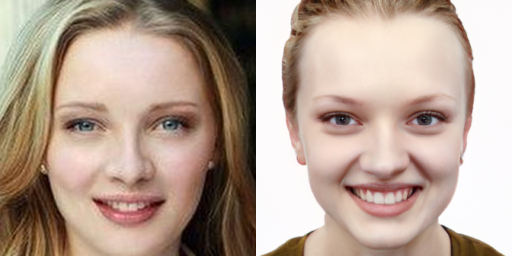}

    \includegraphics[width=0.43\linewidth]{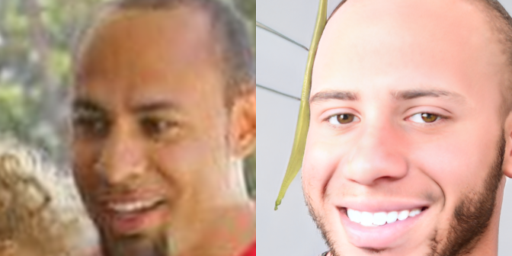}    \hspace{0.5cm}
     \includegraphics[width=0.43\linewidth]{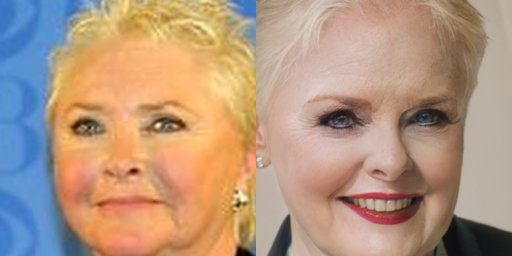}

    \includegraphics[width=0.43\linewidth]{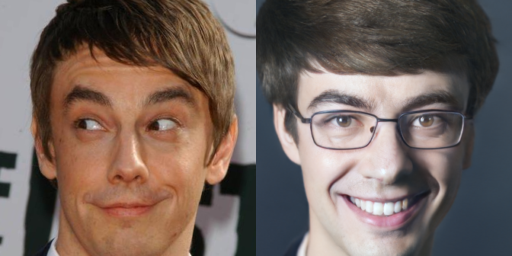}    \hspace{0.5cm}
     \includegraphics[width=0.43\linewidth]{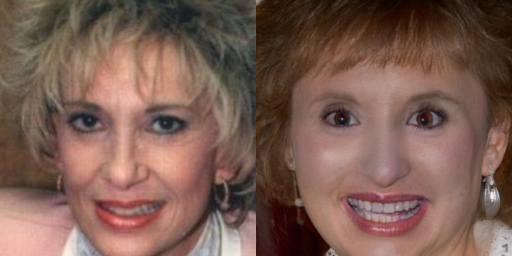}

    \includegraphics[width=0.43\linewidth]{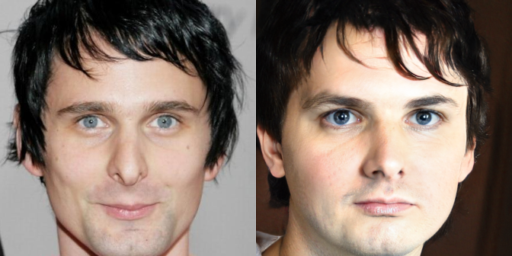}    \hspace{0.5cm}
     \includegraphics[width=0.43\linewidth]{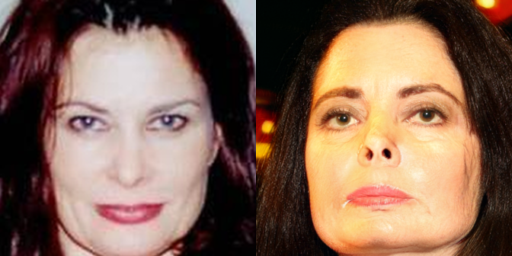}

    \includegraphics[width=0.43\linewidth]{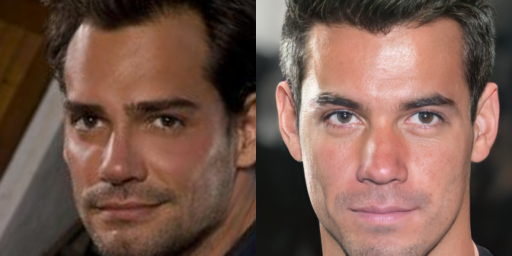}    \hspace{0.5cm}
     \includegraphics[width=0.43\linewidth]{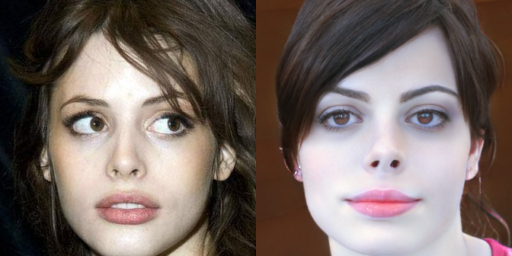}

    \includegraphics[width=0.43\linewidth]{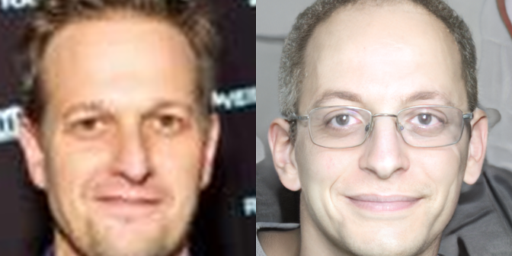}    \hspace{0.5cm}
     \includegraphics[width=0.43\linewidth]{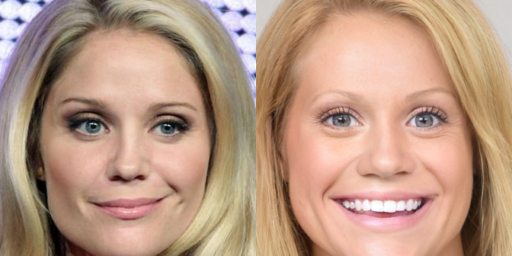}

    \includegraphics[width=0.43\linewidth]{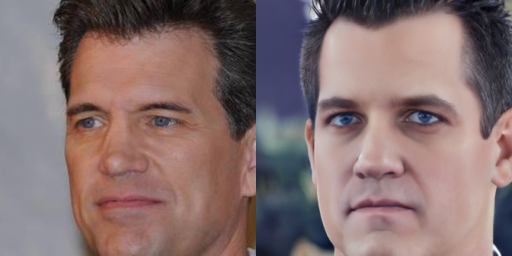}    \hspace{0.5cm}
     \includegraphics[width=0.43\linewidth]{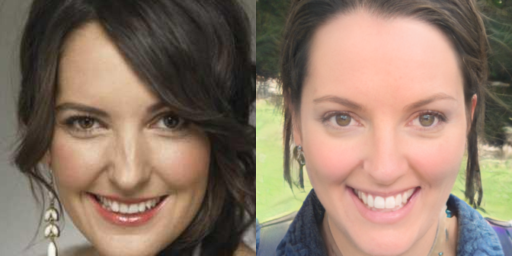}

    \includegraphics[width=0.43\linewidth]{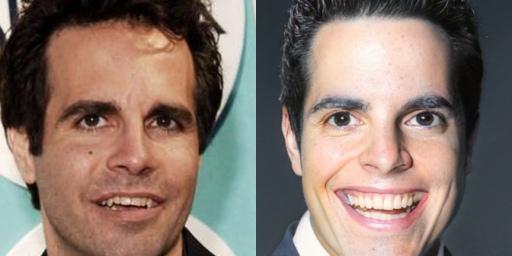}    \hspace{0.5cm}
     \includegraphics[width=0.43\linewidth]{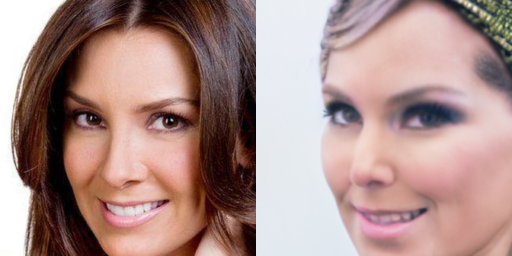}

    \includegraphics[width=0.43\linewidth]{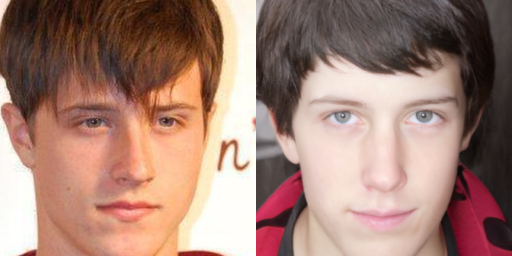}    \hspace{0.5cm}
     \includegraphics[width=0.43\linewidth]{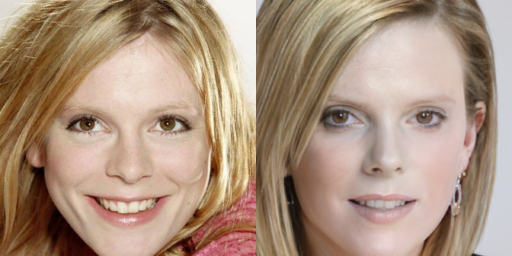}

    \includegraphics[width=0.43\linewidth]{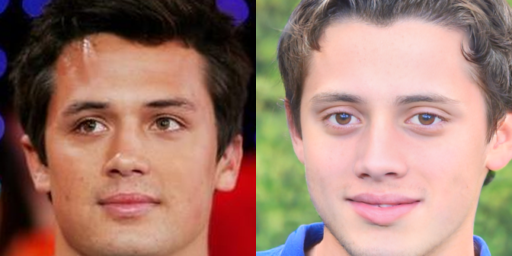}    \hspace{0.5cm}
     \includegraphics[width=0.43\linewidth]{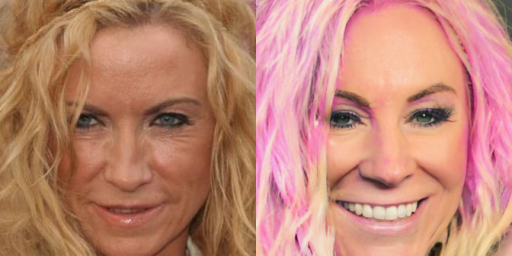}

    \caption{Qualitative results on CelebA dataset using the SMI-AW and $\mathcal{L}_{LOM}$, $M$ = LLaVA-v1.6-7B. For each pair, the left column shows images from the private training dataset, while the right column presents the reconstructed images corresponding to each individual in the left column. }
    \label{fig:celeba_llava}
\end{figure}

\begin{figure}[t!]
    \centering
    \includegraphics[width=0.43\linewidth]{figures_supp/x_recons_priv.png}    \hspace{0.5cm}
    \includegraphics[width=0.43\linewidth]
    {figures_supp/x_recons_priv.png} 

    \includegraphics[width=0.43\linewidth]{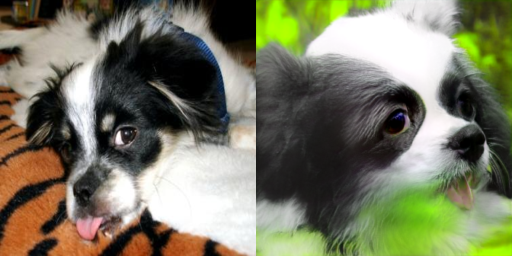}    \hspace{0.5cm}
    \includegraphics[width=0.43\linewidth]
    {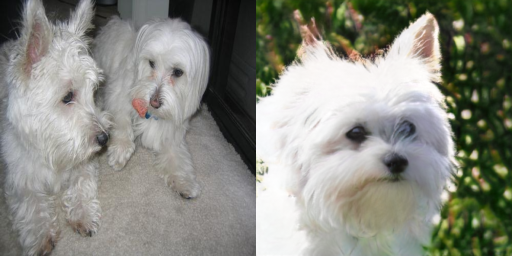}

    \includegraphics[width=0.43\linewidth]{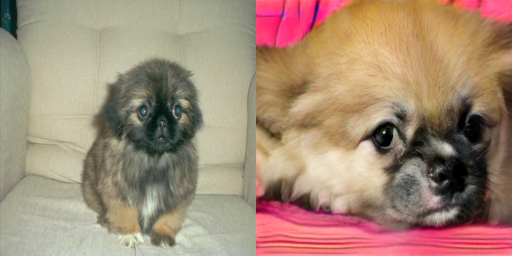}    \hspace{0.5cm}
    \includegraphics[width=0.43\linewidth]
    {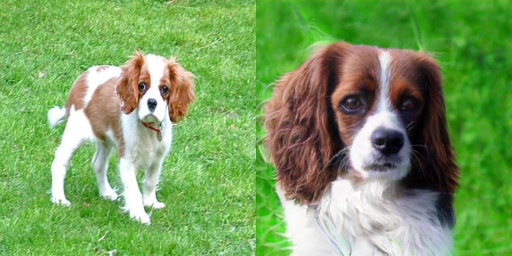}

    \includegraphics[width=0.43\linewidth]{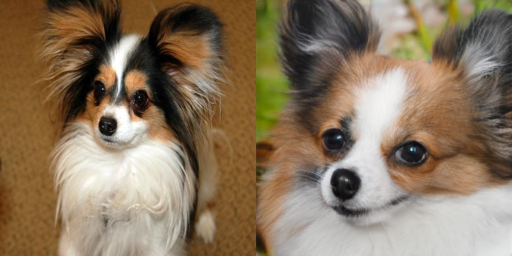}    \hspace{0.5cm}
    \includegraphics[width=0.43\linewidth]
    {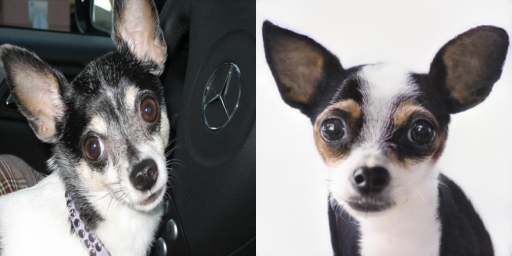}

    \includegraphics[width=0.43\linewidth]{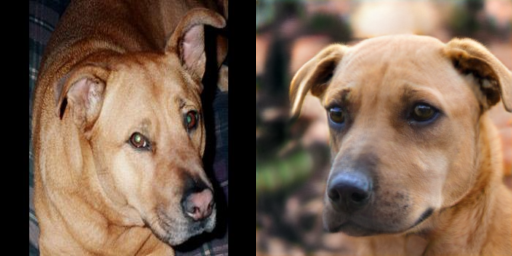}    \hspace{0.5cm}
    \includegraphics[width=0.43\linewidth]
    {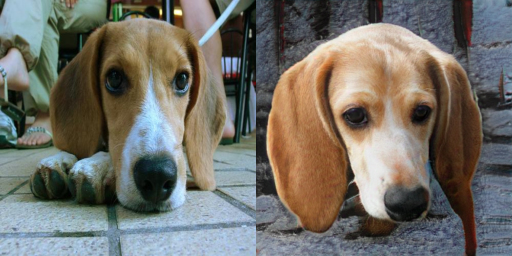}

    \includegraphics[width=0.43\linewidth]{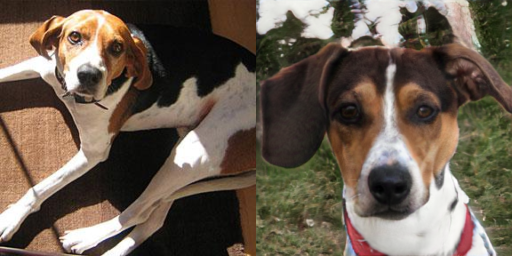}    \hspace{0.5cm}
    \includegraphics[width=0.43\linewidth]
    {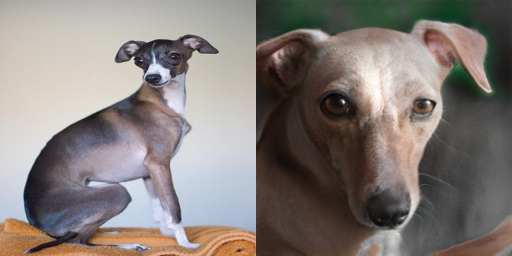}

    \caption{Qualitative results on the Stanford Dogs dataset using the SMI-AW and $\mathcal{L}_{LOM}$, $M$ = LLaVA-v1.6-7B. For each pair, the left column shows images from the private training dataset, while the right column presents the reconstructed images corresponding to each dog breed in the left column. }
    \label{fig:dog_llava}
\end{figure}

Reconstructed images from the FaceScrub dataset using four VLMs, LLaVA-v1.6-7B, MiniGPT-v2,  Qwen2.5-VL, and InternVL2.5 are shown in Figure~\ref{fig:facescrub_llava}, Figure~\ref{fig:facescrub_mingpt},  Figure~\ref{fig:facescrub_qwen}, and Figure~\ref{fig:facescrub_internvl}, respectively.
For the CelebA and Stanford Dogs datasets, reconstructed images using LLaVA-v1.6-7B are presented in Figure~\ref{fig:celeba_llava} and Figure~\ref{fig:dog_llava}.
All reconstructions are generated using SMI-AW with the logit maximization loss $\mathcal{L}_{LOM}$.

For each pair, the left column shows images from the private training dataset, while the right column presents the reconstructed images corresponding to each individual in the left column. 
Qualitative results demonstrate the effectiveness of our method. This strong visual similarity highlights the ability of our model inversion approach to recover identifiable features from the training data.

\subsection{Additional Attention Map Analysis}

\begin{figure}[ht!]
    \centering
\includegraphics[width=0.99\linewidth]{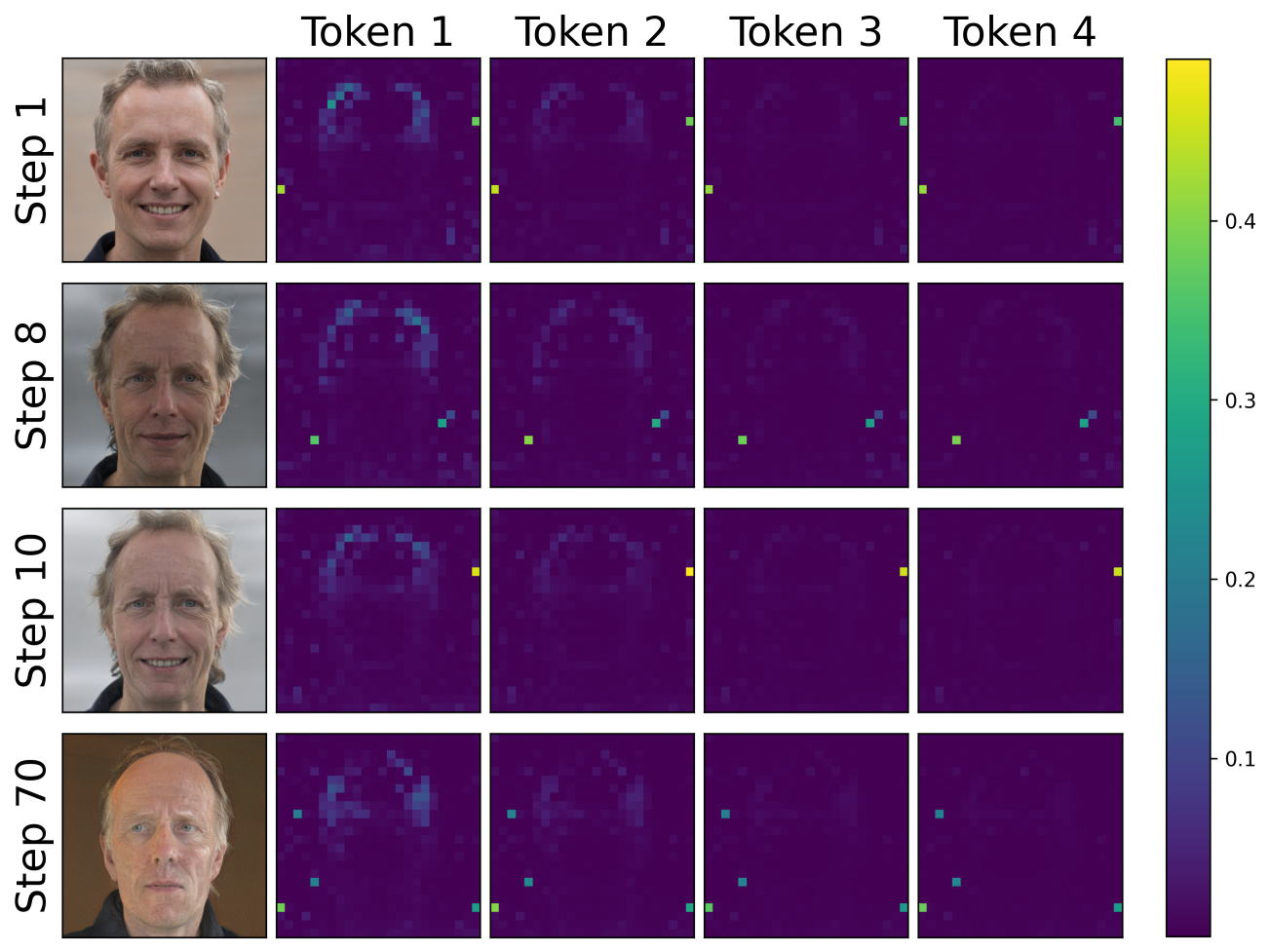} 
    \includegraphics[width=0.87\linewidth]{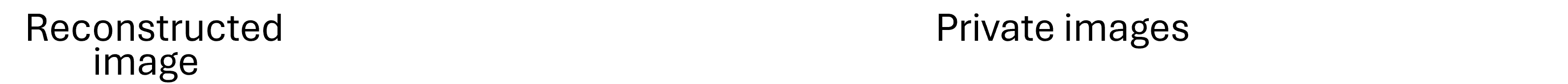}
    \includegraphics[width=0.87\linewidth]{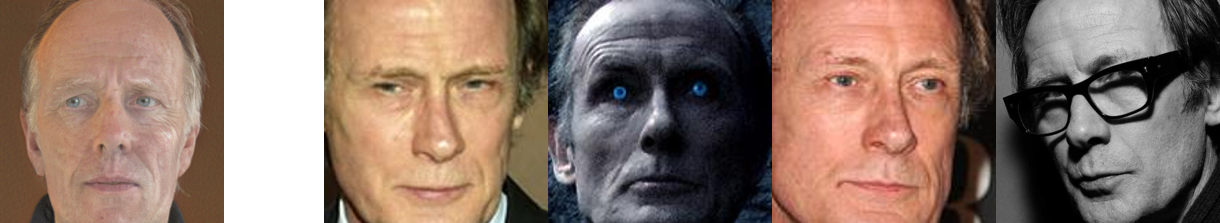}
\rule{1.0\linewidth}{0.4pt}\\   
 \includegraphics[width=0.99\linewidth]{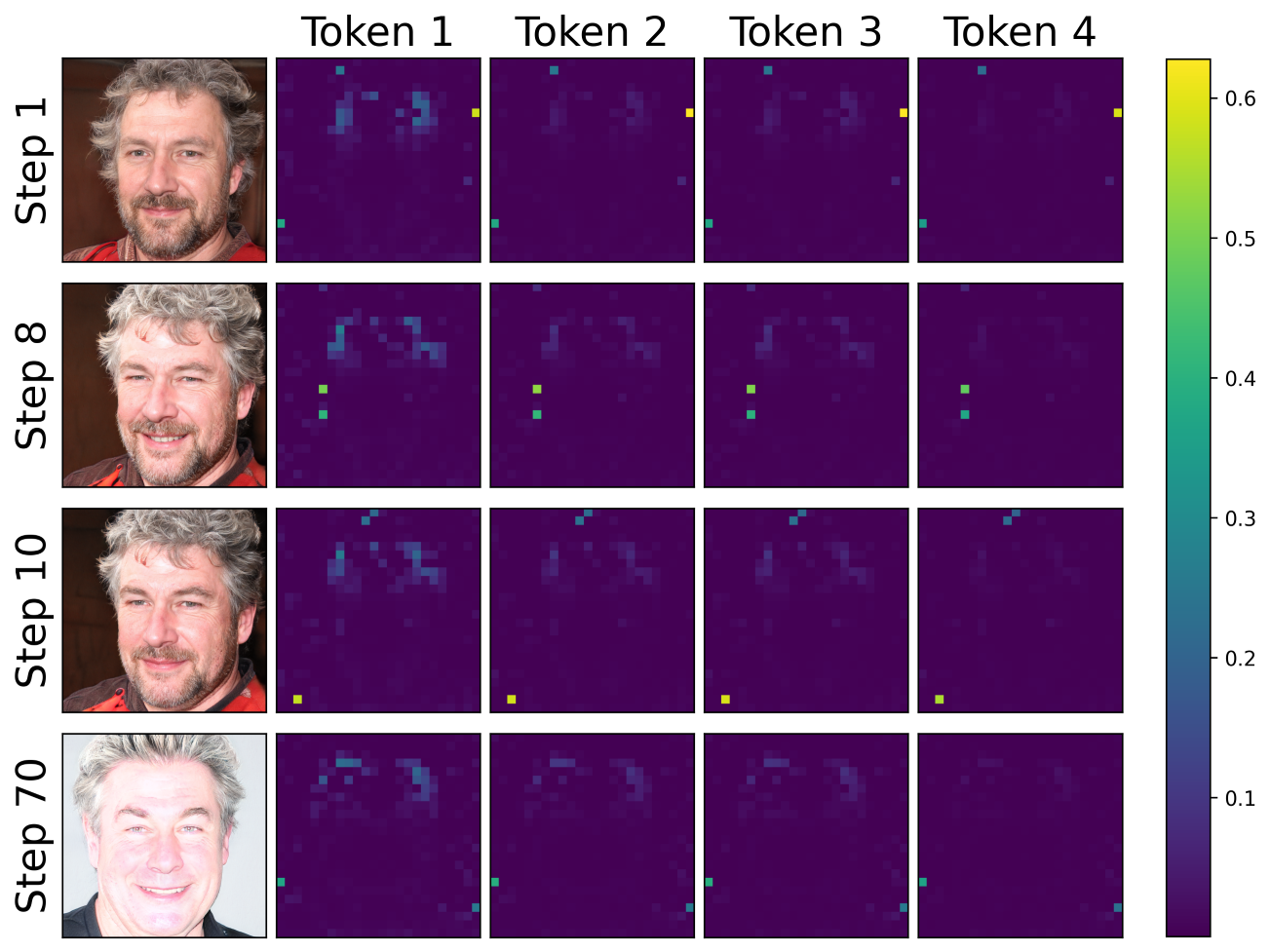} 
    \includegraphics[width=0.87\linewidth]{figures_supp/attention_map/title.png}
    \includegraphics[width=0.87\linewidth]{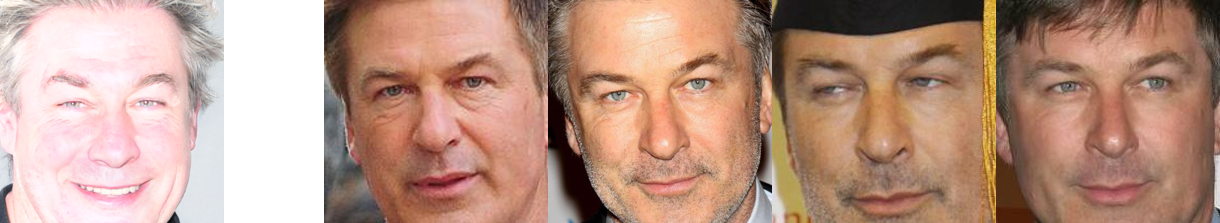}
    \caption{
{\bf Analysis of visual–textual attention across output tokens and inversion steps of LLaVa-1.6-7B model.} 
We visualize the cross-attention map between the reconstructed image and each output token during inversion. 
{\bf Our analysis confirms that token-level gradients vary
substantially in visual informativeness both across tokens and
over time, and this motivates our SMI-AW method with dynamic reweighing.}
    }
    \label{fig:attentionMap_llava}
\end{figure}

\begin{figure}[ht!]
    \centering    
 \includegraphics[width=0.99\linewidth]{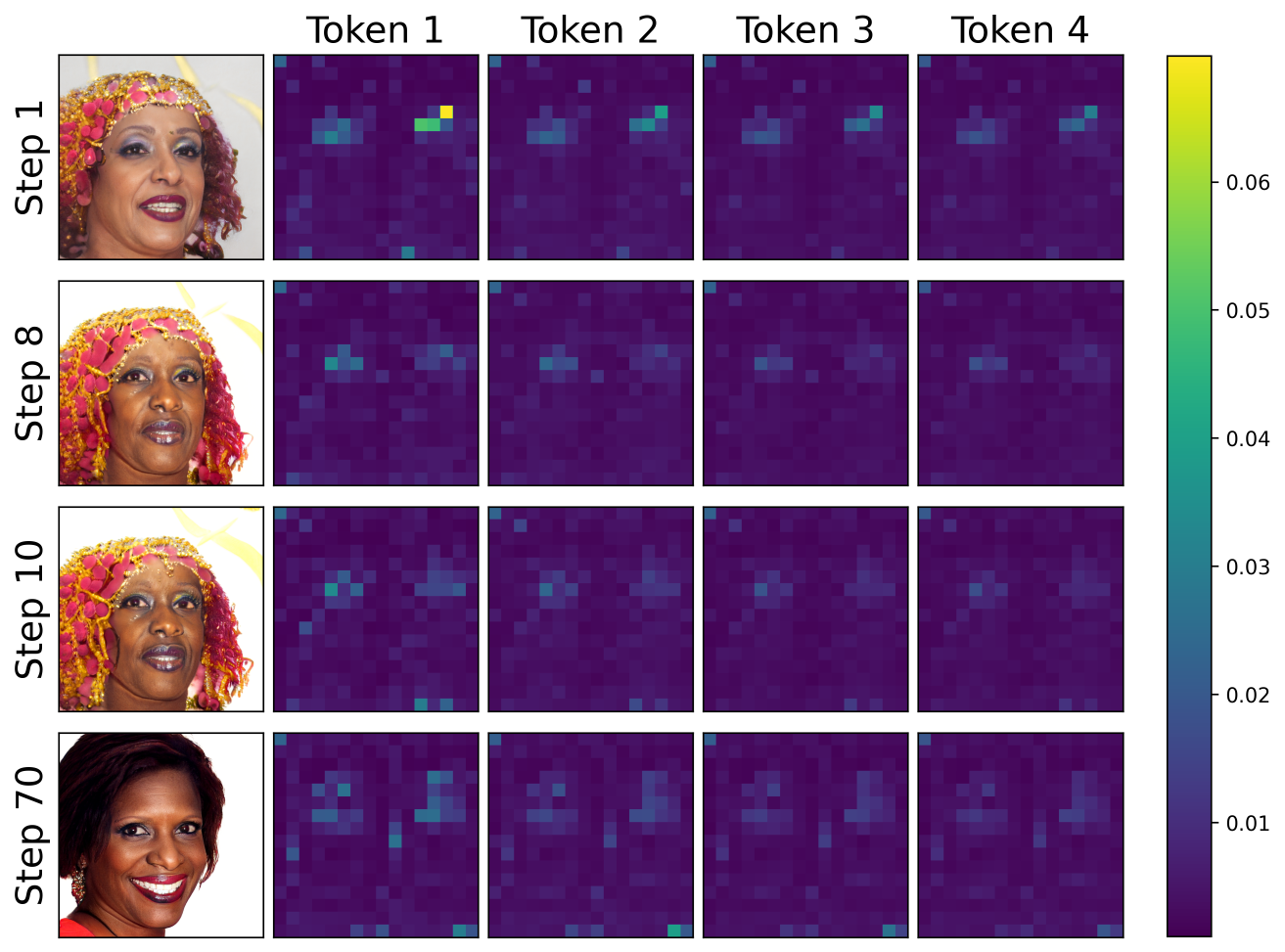} 
    \includegraphics[width=0.87\linewidth]{figures_supp/attention_map/title.png}
    \includegraphics[width=0.87\linewidth]{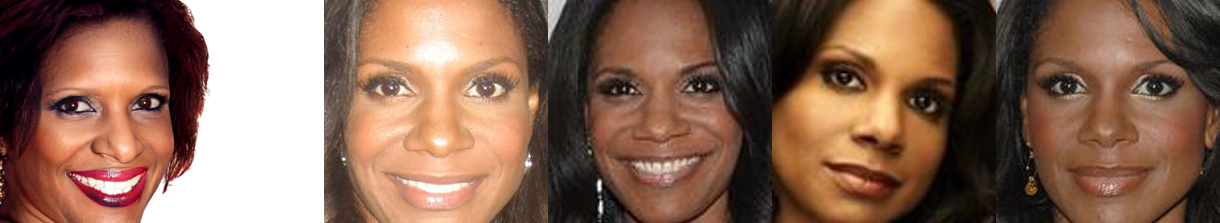}
\rule{1.0\linewidth}{0.4pt}\\   
 \includegraphics[width=0.99\linewidth]{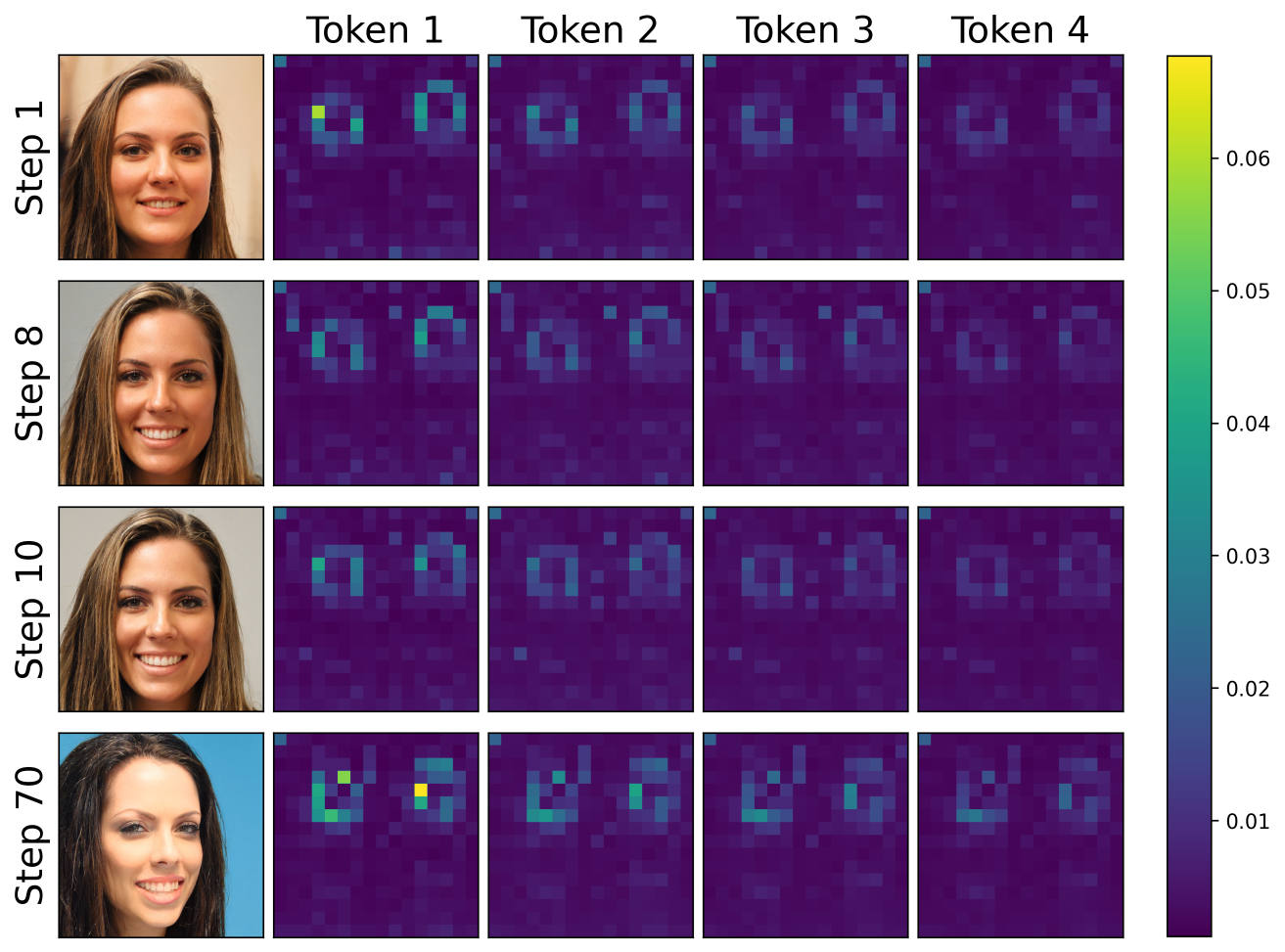} 
    \includegraphics[width=0.87\linewidth]{figures_supp/attention_map/title.png}
    \includegraphics[width=0.87\linewidth]{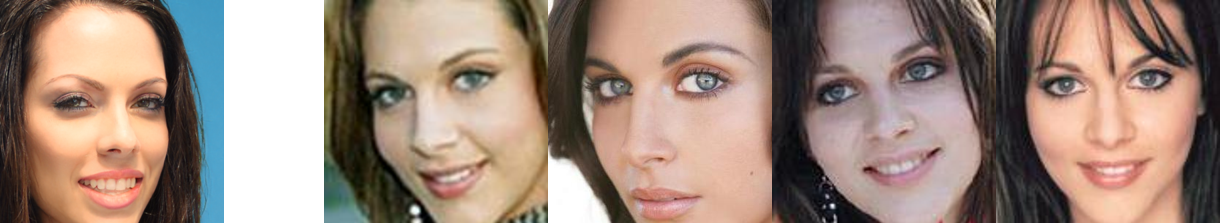}
    \caption{
{\bf Analysis of visual–textual attention across output tokens and inversion steps of MiniGPTv2 model.} 
We visualize the cross-attention map between the reconstructed image and each output token during inversion. 
{\bf Our analysis confirms that token-level gradients vary
substantially in visual informativeness both across tokens and
over time, and this motivates our SMI-AW method with dynamic reweighing.}
    }
    \label{fig:attentionMap_minigpt}
\end{figure}

\begin{figure}[ht!]
    \centering

\includegraphics[width=0.99\linewidth]{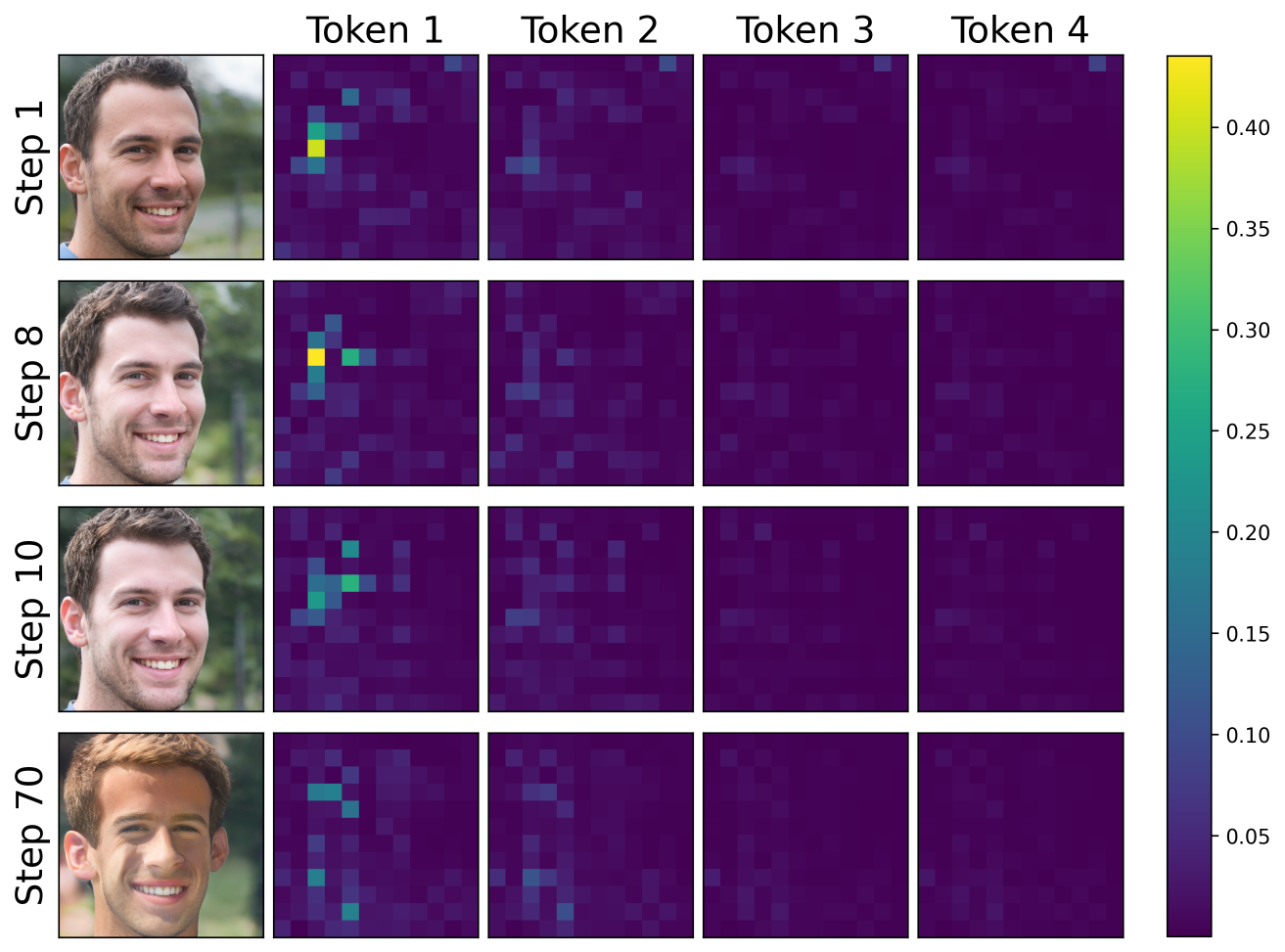} 
    \includegraphics[width=0.87\linewidth]{figures_supp/attention_map/title.png}
    \includegraphics[width=0.87\linewidth]{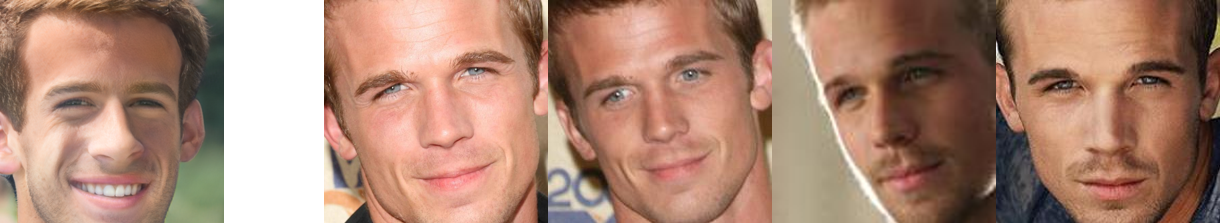}
\rule{1.0\linewidth}{0.4pt}\\   
 \includegraphics[width=0.99\linewidth]{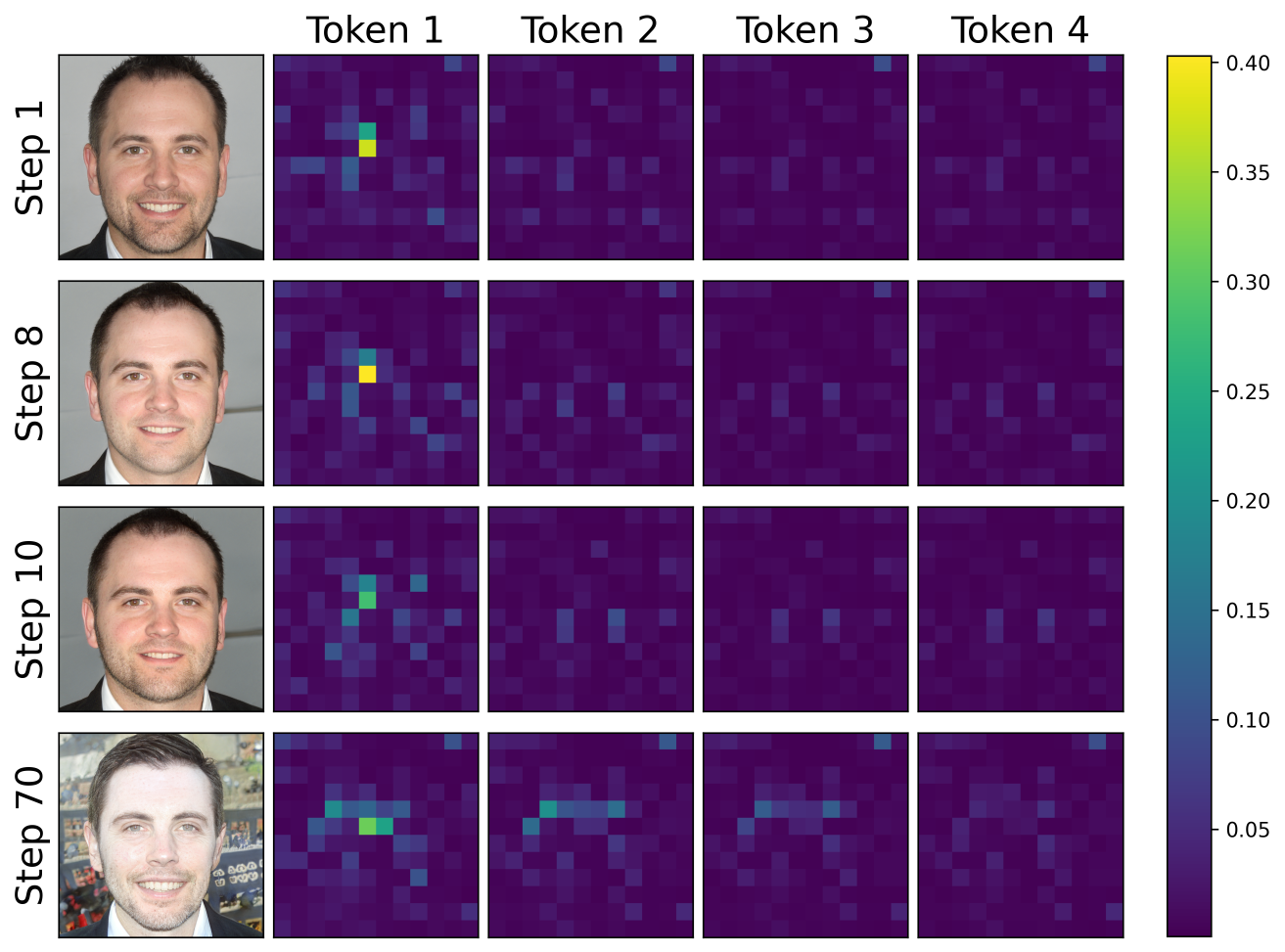} 
    \includegraphics[width=0.87\linewidth]{figures_supp/attention_map/title.png}
    \includegraphics[width=0.87\linewidth]{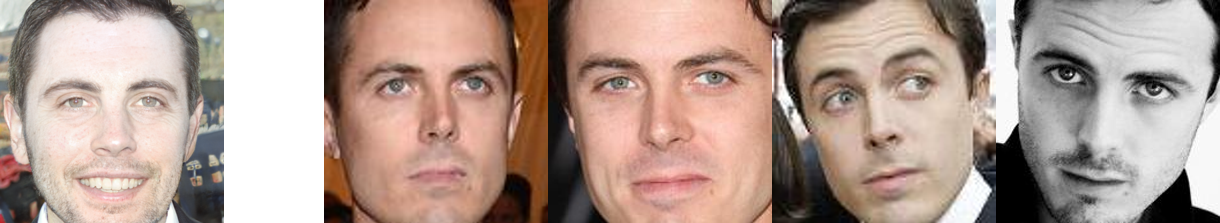}
    \caption{
{\bf Analysis of visual–textual attention across output tokens and inversion steps of Qwen2.5-VL-7B model.} 
We visualize the cross-attention map between the reconstructed image and each output token during inversion.
{\bf Our analysis confirms that token-level gradients vary
substantially in visual informativeness both across tokens and
over time, and this motivates our SMI-AW method with dynamic reweighing.}
    }
    \label{fig:attentionMap_qwen}
\end{figure}

\begin{figure}[ht!]
    \centering
\includegraphics[width=0.99\linewidth]{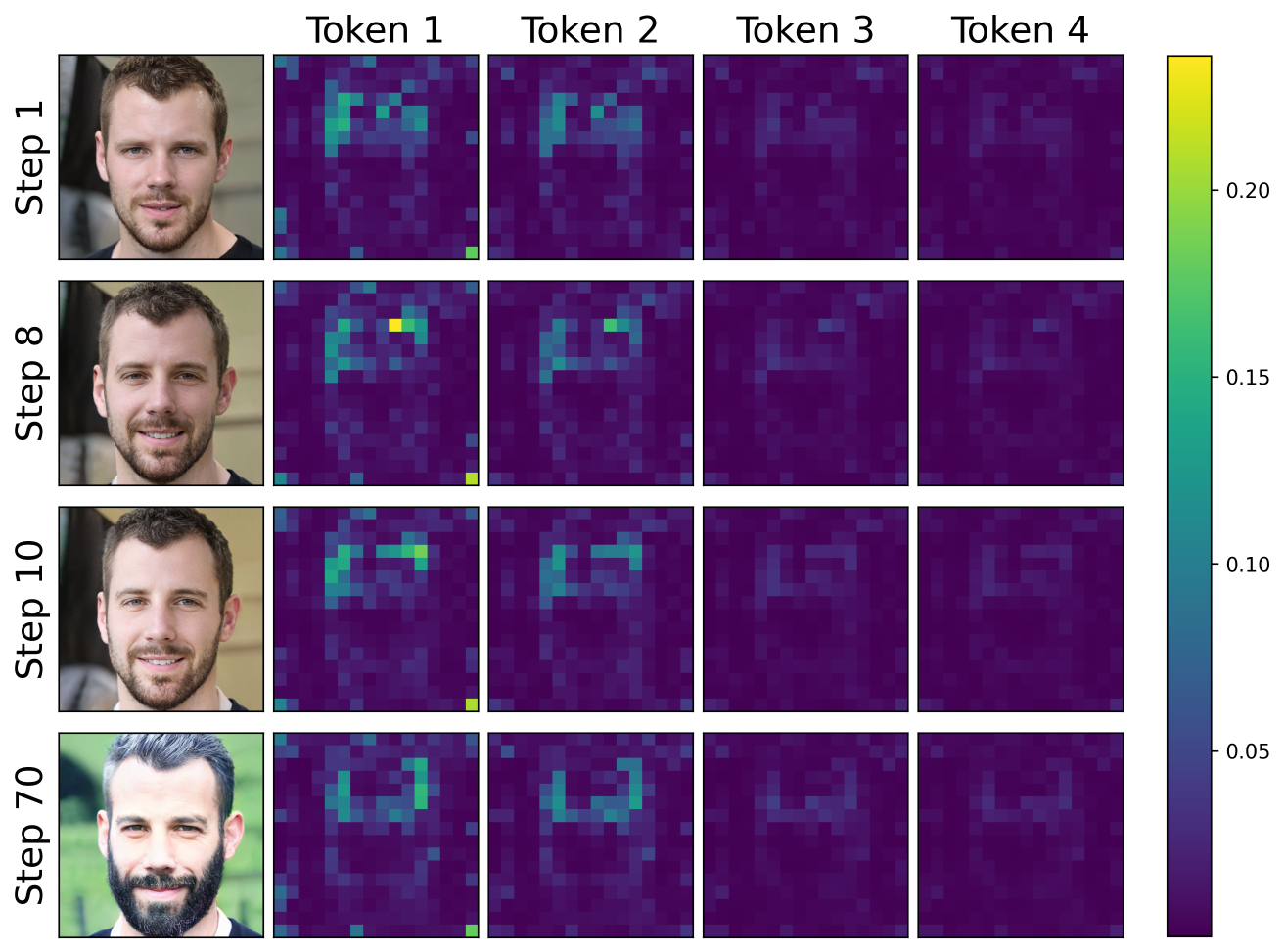} 
    \includegraphics[width=0.87\linewidth]{figures_supp/attention_map/title.png}
    \includegraphics[width=0.87\linewidth]{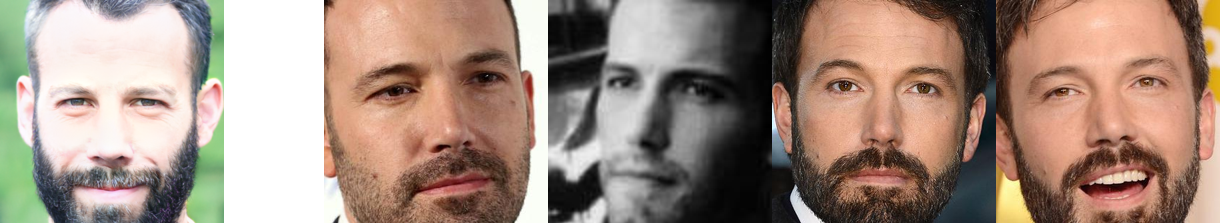}
\rule{1.0\linewidth}{0.4pt}\\   
 \includegraphics[width=0.99\linewidth]{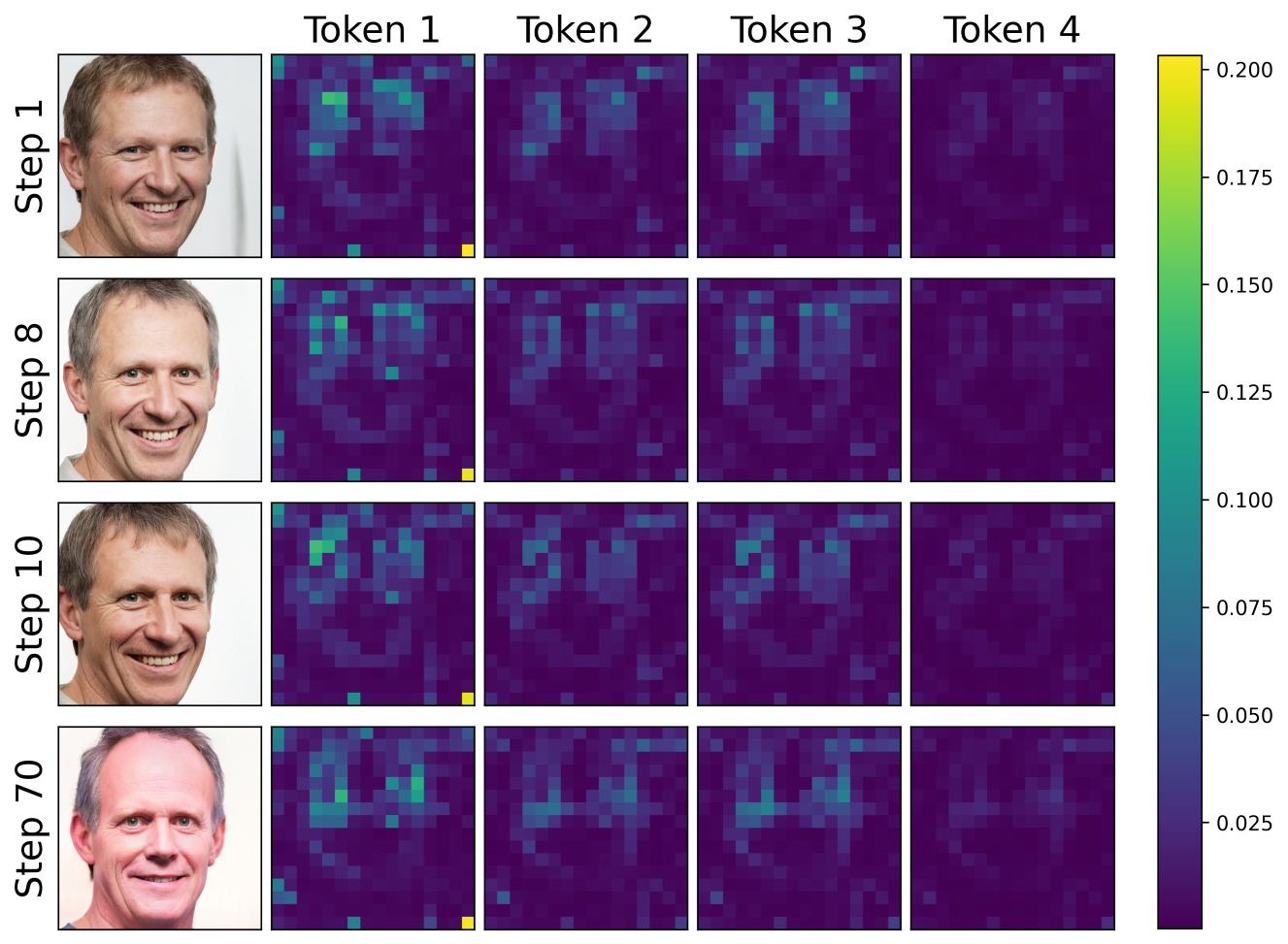} 
    \includegraphics[width=0.87\linewidth]{figures_supp/attention_map/title.png}
    \includegraphics[width=0.87\linewidth]{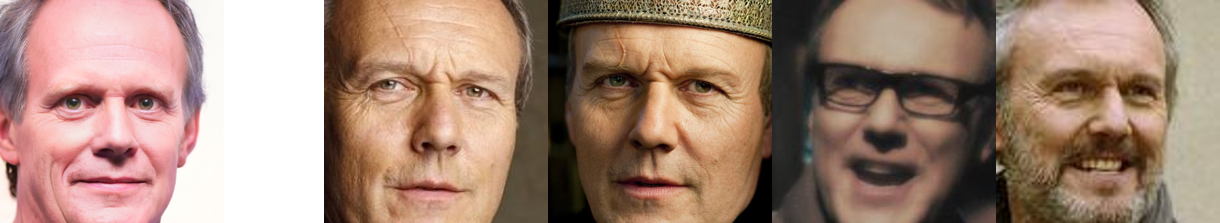}
    \caption{
{\bf Analysis of visual–textual attention across output tokens and inversion steps of InternVL2.5 model.} 
We visualize the cross-attention map between the reconstructed image and each output token during inversion.
{\bf Our analysis confirms that token-level gradients vary
substantially in visual informativeness both across tokens and
over time, and this motivates our SMI-AW method with dynamic reweighing.}
    }
    \label{fig:attentionMap_internvl}
\end{figure}

Additional attention map of four models including LLaVa-1.6-7B, MiniGPTv2, Qwen2.5-VL, and InternVL2.5
are visualized in Figure \ref{fig:attentionMap_llava}, Figure \ref{fig:attentionMap_minigpt}, Figure \ref{fig:attentionMap_qwen}, and Figure \ref{fig:attentionMap_internvl}.
We visualize the cross-attention map between the reconstructed image and each output token during inversion. 
Different tokens exhibit markedly different attention maps: visually grounded tokens show strong attention, while others produce weak responses, indicating limited reliance on the image. Moreover, attention patterns evolve over inversion steps, as a token's dependence on visual input changes when the reconstructed image becomes more consistent with the target output. These observations reveal that token-level gradients vary substantially in visual informativeness both across tokens and over time. This motivates our SMI-AW method, which dynamically reweights token contributions based on their visual attention strength.

\section{Ablation Study}

\subsection{Ablation Study on input prompt $y$}

In this section, we further evaluate SMI-AW using a more diverse set of input prompts $y$. The results are summarized in Table~\ref{tab:input_prompt}. It shows that SMI-AW maintains consistently strong attack performance across different prompt choices, demonstrating its robustness to prompt variation.

\begin{table*}[ht!]
\centering
\caption{ We evaluate SMI-AW using a more diverse set of input prompts $y$.
Here, we use $M =$  LLaVa-v1.6-7B, $\mathcal{D}_{priv} =$ Facescrub and logit maximization loss $\mathbf{L}_{LOM}$.
}
\begin{tabular}{lccccc}
 \hline
\multirow{2}{*}{Input question} & \multirow{2}{*}{$AttAcc_{M}\uparrow$} & \multicolumn{2}{c}{$AttAcc_{D} \uparrow$} &  \multirow{2}{*}{$\delta_{face}\downarrow$} &  \multirow{2}{*}{$\delta_{eval}\downarrow$} \\ \cline{3-4}
 &  & \multicolumn{1}{c}{$Top 1$} & \multicolumn{1}{c}{$Top 5$} &  &   \\ \hline
Who is the person in the image? &  61.01\% & 37.62\%	& 66.16\%	& 0.7265	& 134.94  \\
 What is the person's name in the image? & 59.08\%  & 37.10\%	& 64.62\% &	0.7318 &	135.28 \\
Who is the man/woman in the photo? & 59.98\% & 37.78\% &	64.25\% &	0.7348	& 135.65 \\ \hline 
\end{tabular}
\label{tab:input_prompt}
\end{table*}





\subsection{Error Bar}

We repeat each experiment three times using different random seeds and report the results in Table~\ref{tab:error_bar}. Specifically, we use $M$ = LLaVA-v1.6-7B, $\mathcal{D}_{priv}$ = Facescrub. The results demonstrate that our attacks have low standard deviation.

\begin{table*}[ht!]
\centering
\caption{
Error bars for our two model inversion strategies SMI and SMI-AW. Each experiment was repeated 3 times, and we report the mean and standard deviation of the attack performance. Here, we use $M =$  LLaVa-v1.6-7B, $\mathcal{D}_{priv} =$ Facescrub. All inversion strategies are combined with logit maximization loss $\mathbf{L}_{LOM}$.
}
\begin{tabular}{lccccc}
 \hline
\multirow{2}{*}{Method} & \multirow{2}{*}{$AttAcc_{M}\uparrow$} & \multicolumn{2}{c}{$AttAcc_{D} \uparrow$} &  \multirow{2}{*}{$\delta_{face}\downarrow$} &  \multirow{2}{*}{$\delta_{eval}\downarrow$} \\ \cline{3-4}
 &  & \multicolumn{1}{c}{$Top 1$} & \multicolumn{1}{c}{$Top 5$} &  &   \\ \hline
SMI & 57.83 $\pm$ 1.18\% & 33.50 $\pm$ 0.19\%	& 	61.56	$\pm$ 0.30\% & 0.7473 $\pm$ 0.0006 &	137.89 $\pm$ 2.62 \\ 
SMI-AW &  59.53 $\pm$	0.93\% & 37.76 $\pm$	0.32\% &	66.18 $\pm$	0.13\%	& 0.7265	 $\pm$ 0.0038 & 134.94 $\pm$	0.64\\ \hline
\end{tabular}
\label{tab:error_bar}
\end{table*}

\section{Experimental setting}
\subsection{Inversion Loss Design for VLMs}

In this section, we present the adaptation of the inversion loss from conventional unimodal MI to VLMs. Specifically, the inversion loss in traditional MI typically consists of two components: $\mathcal{L}_{inv} = \mathcal{L}_{id} + \mathcal{L}_{prior}$, where the identity loss $\mathcal{L}_{id}$ guides the generator $G(w)$ to produce images that induce the label $y$ from the target model $M_{DNN}$, and $\mathcal{L}_{prior}$ is a regularization or prior loss.
To extend this to VLMs, we focus on adapting the identity loss $\mathcal{L}_{id}$. We categorize it into two main types: cross-entropy-based and logit-based losses.

\textbf{Cross-entropy-based.} This loss is widely used in MI attacks \citep{zhang2020secret,chen2021knowledge,qiu2024closer} to optimize $w$ such that the reconstruction has the highest likelihood for the target class under the model $M$. For VLMs, we adapt the cross-entropy loss $\mathcal{L}_{CE}$ for each target token $y_i$ as follows:
\begin{equation} 
    \mathcal{L}_{CE}(M(\bt,G(w),y_{<i}),y_i)= -\log \mathbb{P}_M(y_i|\bt,G(w),y_{<i})
    \label{eq:CE}
\end{equation}
$\mathbb{P}_M(y_i|\bt,G(w),y_{<i})$ denotes the predicted probability of token $y_i$, computed over the tokenizer vocabulary of the VLM (e.g., LLaVa-v1.6 uses a vocabulary of 32,000 tokens).

\textbf{Logit-based.} Prior work shows that using cross-entropy loss in MI can lead to gradient vanishing \citep{yuan2023pseudo} or sub-optimal results \citep{nguyen2023re}. To address this, \citet{yuan2023pseudo} and \citet{nguyen2023re} propose optimizing losses directly over logits of a target class. We adopt two such logit-based losses for VLMs: the Max-Margin Loss $\mathcal{L}_{MML}$ \citep{yuan2023pseudo} and the Logit-Maximization Loss $\mathcal{L}_{LOM}$ \citep{nguyen2023re} for a target token $y_i$:
\begin{equation}
\begin{aligned}
     \mathcal{L}_{MML}(M(\bt, G(w), y_{<i}), &y_i) = - l_{y_{i}}(\bt, G(w), y_{<i}) \\&+ \max_{k \neq y_i} l_{k}(\bt, G(w), y_{<i})
\end{aligned}
\label{eq:MML}
\end{equation}
\begin{equation}
\begin{aligned}
 \mathcal{L}_{LOM}(M(\bt,G(w),y_{<i}),y_i) = &-l_{y_i}(\bt,G(w),y_{<i}) \\ &+ \lambda \|f_{y_i} - f_{reg} \|^2_2  
\end{aligned}
\label{eq:LOM}
\end{equation}
Here, $l_{y_i}$ is the logit corresponding to the target token $y_i$, $\lambda$ is a hyperparameter, $f_{y_i} = M^{pen}(\bt,G(w),y_{<i})$ where $M^{pen}()$ denotes the function that extracts the penultimate layer representations for a given input, and 
$f_{reg}$ is a sample activation from the penultimate layer $M^{pen}()$ computed using public images from $\mathcal{D}_{pub}$. Following \citep{nguyen2023re}, the distribution of $f_{reg}$ is estimated over 2000 input pairs $(\bt, \bx_{pub})$, where $\bx_{pub} \in \mathcal{D}_{pub}$. 
$\mathcal{L}_{MML}$ maximizes the logit of the correct token $y_i$ while penalizing the highest incorrect logit to mitigate gradient vanishing. On the other hand, $\mathcal{L}_{LOM}$ also maximizes the correct token's logit to avoid sub-optimality, while additionally penalizing deviations in the penultimate activations to prevent unbounded logits problem.  

\subsection{Evaluation metrics}


\begin{figure}[t!]
    \centering
    \includegraphics[width=1.0\linewidth]{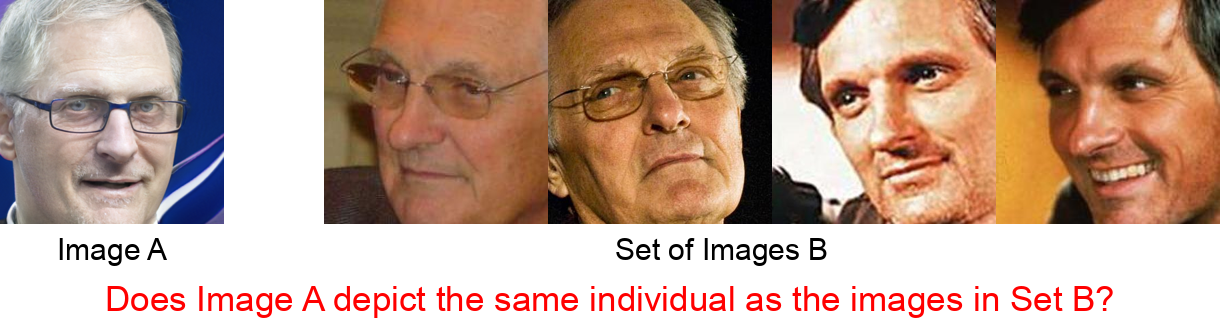}
    \captionof{figure}{An example evaluation query in $\mathcal{F}_{MLLM}$ and human evaluation involves determining whether ``Image A'' depicts the same individual as those in ``Image B.'' ``Image A'' is a reconstructed image of a target textual answer $y$, while ``Image B'' contains four real images of the same target textual answer $y$. Gemini or human evaluators respond with ``Yes'' or ``No'' to indicate whether ``Image A'' matches the identity shown in ``Image B.''
}
    \label{fig:user_study}
\end{figure}

In this section, we provide a detailed implementation for five metrics used in our work to access MI attacks.

\begin{itemize}
    \item \textbf{Attack accuracy.} Attack accuracy measures the success rates of MI attacks. Following existing literature, we compute attack accuracy via three frameworks:
    
    \begin{itemize}
        \item \textbf{Attack accuracy evaluated by conventional evaluation framework $\mathcal{F}_{DNN}$ ($AttAcc_{D} \uparrow$)} \citep{zhang2020secret,chen2021knowledge,struppek2022plug,nguyen2023re,qiu2024closer}. Following \citep{struppek2022plug,struppekcareful}, we use InceptionNet-v3 \citep{szegedy2016rethinking} as the evaluation model. For a fair comparison, we use the identical checkpoints of InceptionNet-v3 for Facescrubs, CelebA and Stanford Dogs from \citep{struppek2022plug} for evaluation of each dataset. We report \textit{Top-1 }and \textit{Top-5} Accuracy. 
        
        \item  \textbf{Attack accuracy evaluated by MLLM-based evaluation framework $\mathcal{F}_{MLLM}$ ($AttAcc_{M} \uparrow$)}. \citep{ho2025revisitingmodelinversionevaluation} demonstrate that $\mathcal{F}_{MLLM}$ can achieve better alignment with human evaluation than $\mathcal{F}_{DNN}$ ($AttAcc_{D} \uparrow$ by mitigating Type-I adversarial transferability. 
        The evaluation involves presenting a reconstructed image (image A) and a set of private reference images (set B) to an MLLM (e.g., Gemini 2.0 Flash), and prompting it with the question: ``Does image A depict the same individual as images in set B?'' If the model responds ``Yes'', the attack is considered successful. An example query is shown in Fig.~\ref{fig:user_study}.
        
        \item \textbf{Attack accuracy evaluated by human $\mathcal{F}_{Human} (AttAcc_{H} \uparrow)$}. Following existing studies \citep{an2022mirror,nguyen2023re}, we conduct the user study on Amazon Mechanical Turk. Participants are asked to evaluate the success of MI-reconstructed by referencing the corresponding private images. Similar to $\mathcal{F}_{MLLM}$, it involves presenting an image A and a set of images B. They are asked to answer “Yes" or “No" to indicate whether image A depicts the same identity as images in set B (see Fig.~\ref{fig:user_study}). Each image pair is shown in a randomized order and displayed for up to 60 seconds. Each user study involves 4,240 participants for the FaceScrub dataset and 8,000 participants for the CelebA dataset.

    \end{itemize}
    \item \textbf{Feature distance.} We compute the $l_2$ distance between the feature representations of the reconstructed and the private training images \citep{struppek2022plug}. Lower values indicate higher similarity and better inversion quality.
    \begin{itemize}
        \item \textbf{$\delta_{eval}$.} Features are extracted by the evaluation model as used in $\mathcal{F}_{DNN}$.
        \item \textbf{$\delta_{face}$.} Features are extracted by a pre-trained FaceNet model \citep{schroff2015facenet}.
    \end{itemize}

\end{itemize}

\subsection{Initial Candidate Selection}
Following the method from \citep{struppek2022plug}, we perform an initial selection to identify promising candidates for inversion. We begin by sampling 2000 latent vectors, denoted as $\{w\}_{i=1}^{2000}$, from the prior distribution. For each $w$, we evaluate the target VLMs loss. We then select the top $n$ vectors with the lowest loss to serve as our initialization candidates. In our experiments, we set $n=16$ to create 16 candidates for attacks.

\subsection{Final Selection}
To select the final reconstructed image, we perform a final selection step, also following the method from \citep{struppek2022plug}. This step aims to identify the reconstructed images that have the highest confidence. For each of the $n$ initialization candidates, we apply 10 random data augmentations and re-evaluate the target VLMs loss. We calculate the average loss for each candidate across these augmentations and select the $n/2$ candidates with the lowest average loss as the final attack outputs.

\section{Related Work}

\textbf{Model Inversion. }Model Inversion (MI) seeks to recover information about a model’s private training data via pretrained model. Given a target model \( M \) trained on a private dataset \( \mathcal{D}_{\text{priv}} \), the adversary aims to infer sensitive information about the data in \( \mathcal{D}_{\text{priv}} \), despite it being inaccessible after training. MI attacks are commonly framed as the task of reconstructing an input that the model \( M \) would classify as belonging to a particular label \( y \). The foundational MI method is introduced in \citep{fredrikson2014privacy}, demonstrating that machine learning models could be exploited to recover patients' genomic and demographic data.

\textbf{Model Inversion in Unimodal Vision Models. } Model Inversion (MI) has been extensively studied to reconstruct private training images in unimodal vision models. For example, in the context of face recognition, MI attacks attempt to recover facial images that the model would likely associate with a specific individual.

Building on the foundational work of \citep{fredrikson2014privacy}, early MI attacks targeting facial recognition are proposed in \citep{fredrikson2015model,yang2019neural}, demonstrating the feasibility of reconstructing recognizable facial images from the outputs of pretrained models. However, performing direct optimization in the high-dimensional image space is challenging due to the large search space. To address this, recent advanced generative-based MI attacks have shifted the search to the latent space of deep generative models \citep{zhang2020secret,wang2021variational,chen2021knowledge,yang2019neural,yuan2023pseudo,nguyen2023re,struppek2022plug,qiu2024closer}. 

Specifically, GMI \citep{zhang2020secret} and PPA \citep{struppek2022plug} employ WGAN \citep{arjovsky2017wasserstein} and StyleGAN \citep{karras2019style}, respectively, trained on an auxiliary public dataset \( \mathcal{D}_{\text{pub}} \) that similar to the private dataset \( \mathcal{D}_{\text{priv}} \). The pretrained GAN is served as prior knowledge for the inversion process. To improve this prior knowledge, KEDMI \citep{chen2021knowledge} trains inversion-specific GANs using knowledge extracted from the target model \( M \). PLGMI \citep{yuan2023pseudo} introduces pseudo-labels to enhance conditional GAN training. IF-GMI \citep{qiu2024closer} utilizes intermediate feature representations from pretrained GAN blocks. Most recently, PPDG-MI \citep{peng2024pseudo} improves the generative prior by fine-tuning GANs on high-quality pseudo-private data, thereby increasing the likelihood of sampling reconstructions close to true private data. Beyond improving GAN-based priors, several studies focus on improving the MI objective including max-margin loss \citep{yuan2023pseudo} and logit loss \citep{nguyen2023re} to better guide the inversion process. Additionally, LOMMA \citep{nguyen2023re} introduces the concept of augmented models to improve the generalizability of MI attacks.

Unlike MI attacks, MI defenses aim to reduce the leakage of private training data while maintaining strong predictive performance. Several approaches have been proposed to defend against MI attacks. MID \citep{wang2021improving} and BiDO \citep{peng2022bilateral} introduce regularization-based defenses that include the term of regularization in the training objective. The crucial drawback of these approaches is that the regularizers often conflict with the training objective resulting in a significant degradation in model's utility. Beyond regularization-based strategies, TL-DMI \citep{ho2024model} leverages transfer learning to improve MI robustness, and LS \citep{struppekcareful} applies Negative Label Smoothing to mitigate inversion risks. Architectural approaches to improve MI robustness have also been explored in \citep{koh2024vulnerability}. More recently, Trap-MID \citep{liu2024trap} introduces a novel defense by embedding trapdoor signals into $M$. These signals act as decoys that mislead MI attacks into reconstructing trapdoor triggers instead of actual private data.

\textbf{Model Inversion in Multimodal Large Vision-Language Models. } Large Vision-Language Models (VLMs) are increasingly deployed in many real-world applications across diverse domains, including sensitive areas
\citep{liu2024llavanext,Qwen2.5-VL,chen2023minigptv2,chen2024internvl,chen2024expanding}. Unlike unimodal vision models, VLMs are designed to process both image and text inputs and generate text responses. A typical VLM architecture includes a text tokenizer to encode textual inputs into text tokens, a vision encoder to extract image features as image tokens, and a lightweight projection layer that maps image tokens into the text token space. These tokens are then concatenated and passed through a LLM to produce the final response. This multimodal processing pipeline fundamentally distinguishes VLMs from traditional unimodal vision models.

As VLMs are being adopted more widely, including in privacy-sensitive scenarios, understanding their potential vulnerability to data leakage via MI attacks becomes critical. {\bf However, while MI attacks have been extensively studied in unimodal vision models, to the best of our knowledge, there has been no prior work investigating MI attacks on multimodal VLMs. To fill this gap, \textit{we conduct the first study on MI attacks targeting VLMs and propose a novel MI attack framework specifically tailored to the multimodal setting of VLMs.
   }}

\section{Discussion}
\subsection{Broader Impacts}

Our work reveals, for the first time, that VLMs are vulnerable to MI attacks. As VLMs are increasingly deployed in many applications including sensitive domains, this poses serious privacy risks. Although our work focuses on developing a new MI attack for VLMs, we also provide a fundamental understanding for the development of MI defenses in multimodal systems. We hope this work encourages the community to incorporate privacy audits in VLM deployment and to pursue principled model design that mitigates data leakage.

Our methods are intended solely for research and defense development. We strongly discourage misuse and emphasize responsible disclosure when evaluating model vulnerabilities.

\subsection{Limitations}
While following conventional MI attacks to focus on facial images and dog breeds, a more diverse domain scenarios, such as natural scenes or medical images, remain an important direction for future research. Moreover, 
evaluations with more models can further support our claims.



\end{document}